\documentclass[lettersize,journal]{IEEEtran}
\usepackage{amsmath,amsfonts}
\usepackage{algorithmic}
\usepackage{algorithm}
\usepackage{array}
\usepackage{doi}
\usepackage[caption=false,font=normalsize,labelfont=sf,textfont=sf]{subfig}
\usepackage{textcomp}
\usepackage{stfloats}
\usepackage{url}
\usepackage{relsize}
\usepackage{verbatim}
\usepackage{graphicx}
\usepackage{cite}
\usepackage{booktabs}
\usepackage{multirow}
\usepackage{threeparttable}
\usepackage{bbding}
\usepackage{pifont}
\usepackage{booktabs,siunitx}
\usepackage{colortbl}
\usepackage{hyperref}[colorlinks,linkcolor=blue]
\hypersetup{citecolor=red}
\graphicspath{./figure}

\begin{document}

\title{{Clothoid Curve-based Emergency-Stopping Path Planning with Adaptive Potential Field for Autonomous Vehicles}}

\author{Pengfei Lin, Ehsan Javanmardi, and~Manabu~Tsukada \vspace{-2em}
}


\maketitle

\begin{abstract}
The Potential Field (PF)-based path planning method is widely adopted for autonomous vehicles (AVs) due to its real-time efficiency and simplicity. PF often creates a rigid road boundary, and while this ensures that the ego vehicle consistently operates within the confines of the road, it also brings a lurking peril in emergency scenarios. If nearby vehicles suddenly switch lanes, the AV has to veer off and brake to evade a collision, leading to the "blind alley" effect. In such a situation, the vehicle can become trapped or confused by the conflicting forces from the obstacle vehicle PF and road boundary PF, often resulting in indecision or erratic behavior, even crashes. 
To address the above-mentioned challenges, this research introduces an Emergency-Stopping Path Planning (ESPP) that incorporates an adaptive PF (APF) and a clothoid curve for urgent evasion. First, we design an emergency triggering estimation to detect the "blind alley" problem by analyzing the PF distribution. Second, we regionalize the driving scene to search the optimal breach point on the road PF and the final stopping point for the vehicle by considering the possible motion range of the obstacle. Finally, we use the optimized clothoid curve to fit these calculated points under vehicle dynamics constraints to generate a smooth emergency avoidance path. The proposed ESPP-based APF method was evaluated by conducting the co-simulation between MATLAB/Simulink and CarSim Simulator in a freeway scene. The simulation results reveal that the proposed method shows increased performance in emergency collision avoidance and renders the vehicle safer, in which the duration of wheel slip is 61.9\% shorter, and the maximum steering angle amplitude is 76.9\% lower than other potential field-based methods.
\end{abstract}


\section{Introduction}
\IEEEPARstart{A}{pproximately} 1.3 million people lose their lives every year due to traffic crashes, and there are 20 to 50 million people suffer non-fatal injuries, with many suffering the perpetual disability and\/or mental handicap \cite{inju2021-rp,Curtin2021-ux}. Therefore, a novel generation of intelligent transportation systems (ITS) with autonomous vehicles (AVs) as the core component is proposed and developed worldwide to reduce casualties in the next decade. Soon after, the manufacturers started to conduct road tests for the AVs to fast commercialization. However, the immature AVs have unfortunately caused a series of traffic tragedies. According to the autonomous vehicle collision reports from the Department of Motor Vehicles (DMV), California \cite{AVreport2021}, there have been approximately 546 traffic collision events related to AVs until January 2023.

Path planning, as a crucial module of autonomous driving systems, is responsible for generating a collision-free trajectory. Typically, path planning is divided into two main categories: global path planning (also known as route planning) and local path planning. Global path planning assumes knowing the environment in advance, such as an available map. And then, graph search-based methods are usually used to obtain the globally shortest path with the given current position and destination \cite{Gonzalez2016-zj, Li2017-dy}. Local path planning refers to real-time generating trajectories in the surrounding environment known locally and can be reconstructed based on the sensors. Function-optimized methods and parametric curves are frequently applied in local path planning due to the requirement of high real-time performance \cite{Claussmann2020-so}.

Potential field (PF), one of the most popular path planning methods, is the function-optimized approach with a well-defined mathematical formulation, a high real-time performance, and a simple structure. The PF was first applied to tackle the path-planning tasks of mobile robots \cite{Khatib1985-ia}. It is originally inspired by classical mechanics, formulating virtual forces from the obstacles and the target point \cite{Pamosoaji2013-rk}. This method assigns an attractive force to the goal position for driving the robot toward its goal while establishing a repulsive force on obstacle vehicles to prevent a collision. Due to the complex traffic conditions, real-time performance and driving safety are two cardinal indicators to evaluate the level of AVs \cite{Katrakazas2015-uw}. Therefore, the PF satisfies the aforementioned requirements and has become one of the mainstream in the current planning algorithms. \cite{Li2017-dy}.

Collision avoidance is the primary task of path planning. Compared to mobile robots, the difficulty of collision avoidance for AVs requires higher standardization, as the nonholonomic constraints of an urban vehicle are more complicated than those of an indoor mobile robot \cite{Samuel2016-xf}. In particular, AVs should handle more significant emergency obstacle avoidance, including unexpected events on the freeway, such as sudden lane-changed obstacles without a pre-warning. Therefore, to address {the ``blind alley'' problem where the AVs can be trapped or indecisive leading to crashes}, {we propose a unique Emergency-Stopping Path Planning (ESPP) method with the clothoid curve and the adaptive potential field (APF)} to enable AVs to achieve safe emergency collision avoidance and stop manoeuvers. The contributions of this study are briefly summarized as follows:
\begin{itemize}
    \setlength{\itemsep}{0pt}
    \setlength{\parsep}{0pt}
    \setlength{\parskip}{0pt}
    \item We introduced a clothoid curve to design the {ESPP}; the clothoidal coefficients are obtained by solving a constrained quadratic programming (QP) problem.
    \item {We propose to open a breach on the road PF based on the local reference waypoints, which can navigate the vehicle for a safe stopping maneuver on the roadside.}
    \item We computed the terminal side of the {ESPP} by considering the road structure and the predicted motion range of the obstacle.
\end{itemize}

\begin{figure*}[t]
    \centering
    \includegraphics[width=\hsize]{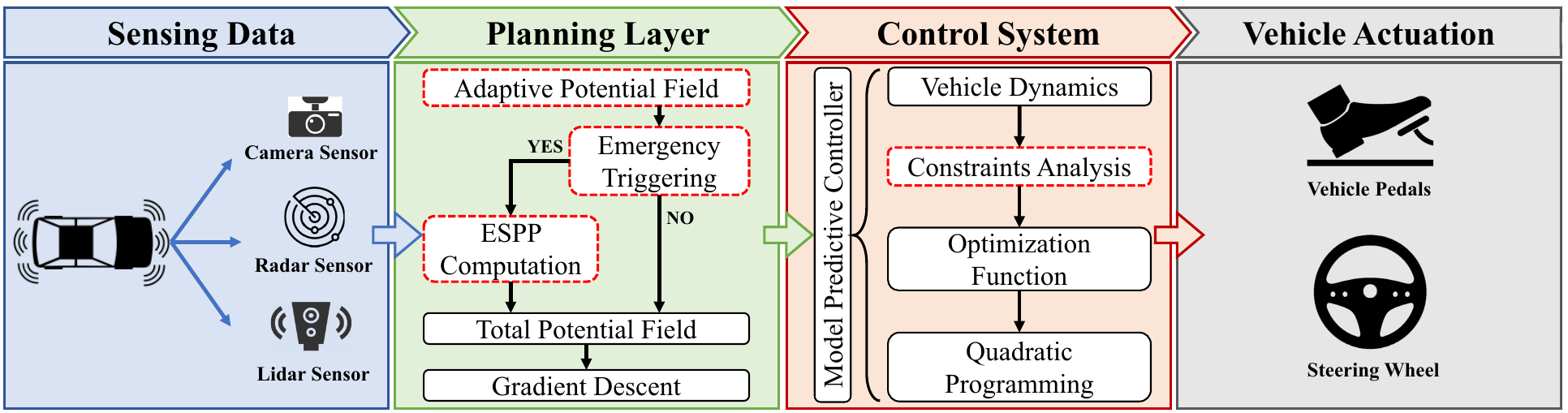}
    \caption{Overall autonomous driving system framework with the embedded {ESPP} method: Sensing data gathers the information collected by sensors; Planning layer monitors the emergency and triggers the {ESPP} computation if the emergency is sensed; Control system outputs the direct signals to Vehicle Actuators for longitudinal and lateral maneuvers}
    \label{sys}
\end{figure*}

The proposed {ESPP} method is embedded mainly in the planning module, including 
the APF, emergency triggering, and {ESPP} computation, as shown in Fig. \ref{sys} (red dotted frames). The complete autonomous driving system framework includes the Sensing, Planning, Control, and Vehicle Actuation. {In addition, we list the following assumptions in this study.}
\begin{itemize}
    \item {We assume that this work mainly concentrates on straight roads.}
    \item {We assume that the collision can not be avoided within the road and an emergency stopping lane or open space is needed.}
    \item {We assume that the ego vehicle performs a total (maximum) braking maneuver when an emergency occurs.}
\end{itemize}

The rest arrangement of this article is as follows: Section II discusses the related work in the past few years. And then, the adaptive potential field is introduced in Section III. Next, Section IV studies the clothoid curve-based {ESPP} method. Section V describes the model predictive controller for the path-tracking task. Then, Section VI illustrates the numerical analysis for validation. Finally, the conclusion and discussion are made in Section VII.

\section{RELATED WORK}\label{rw}

Ji et al. \cite{Ji2017-xj} combined the PF with a specific multi-constrained MPC to tackle path planning and tracking tasks for AVs. However, the generated path did not always conform to vehicle dynamics. Moreover, they only studied a single forward obstacle with constant speed or acceleration, {no emergencies were considered}. Rasekhipour et al. \cite{Rasekhipour2017-yh} used the signed distance (SD) to calculate the minimum distance between the ego vehicle and obstacles that were applied in the potential functions; the scenarios were focused on vehicle merging problems, and the local minima were not considered{, which was unable to deal with urgent events}. {Li et al. and Wang et al. \cite{Li2021-ko, Wang2015-yi} designed a novel driving safety field that contains a PF, a kinetic field, and a behavior field to consider the static and moving objects and pedestrians, as well as the individual characteristics of human drivers. However, the safety driving field model was used as a threat-assessment strategy, which had a finite discussion on the path generation, including the path quality, determination of the model parameters, etc. Wang et al. \cite{Wang2019-cv, Wang2020-va} proposed to include the crash severity and artificial PF into the objective function of an MPC to achieve conventional collision avoidance with the lowest crash mitigation. However, the case studies assumed that collisions were unavoidable. Similarly, Lin et al. \cite{Lin2020-rb, Lin_undated-gd} studied the integration of the PF with a clothoid curve for collision avoidance in a waypoint tracking scenario; however, they assumed that the obstacles were driven only in a straight lane without lane-changing behavior{, lacking emergency analysis}. Lu et al. \cite{Lu2020-fm} proposed an improved APF to adapt both the acceleration/deceleration and the mass of the obstacle to a potential function by using two Gaussian-like functions on both curve and straight roads. {However, the local minima were ignored by assuming that the obstacles' motions were pre-known and did not discuss emergency scenarios.}

Recently, Wang et al. \cite{Wang2021-xs} proposed a PF-based path planning that is adjusted to a kinematic vehicle model for curvy roads, but the simulation was evaluated under the assumption of constant speed, and the obstacles are either stationary or traveling in a straight line at a constant low speed. {To solve the zigzag path caused by traditional PF, Li et al. \cite{Li2022-sz} proposed an optimization-based path planning that used a dynamic enhanced firework algorithm (dynEFWA)-APF. However, the road PF was set to be inviolable, and the speed was also assumed constant. Lin et al. \cite{Lin2022-op, Lin2022-st} presented a unique safe tunnel-based model predictive path-planning controller (STMPC) with the APF to solve the local minima problem that appeared on the highway; however, the longitudinal speed is assumed to be constant, and the road PF cannot be adjusted. To tackle the overtaking problem, Xie et al. \cite{Xie2022-jm} presented a distributed motion planning framework via artificial PF to introduce the notion of velocity difference PF and acceleration difference PF for vehicle platoons. However, the road PF remained unchangeable, and local minima weren't taken into account. Similarly, Wu et al. \cite{Wu2022-ae} proposed a human-like motion planning algorithm for expressway lane-changing behavior that used the artificial PF to analyze the coupling relationship, but the study did not discuss the local minima and assumed the PF is always working. Ji et al. \cite{Ji2023-gj} proposed a three-dimensional PF (TriPField) that combines the ellipsoid PF with a Gaussian velocity field (GVF) to conquer local minima by considering the road user's geometric shape. However, road PF was not discussed because it assumes collision avoidance can be completed within the road. To improve the tracking accuracy, Chu et al. \cite{Chu2022-tg} used the artificial PF to compute the reference trajectory and combined the MPC with PID feedback for the tracking task. Still, this study established an unalterable road PF and focused more on tracking performance. Shang et al. \cite{Shang2023-ye} presented a novel artificial PF that has the flexibility to fit different shapes of road structures. Particularly the proposed PF was implemented in an MPC controller with collision mitigation. However, a tremendous potential value was allocated to the road edges, which means the road edge is also unbreakable. To further improve the motion planning and tracking performance, Du et al. \cite{Du2023-xk} developed global heuristic planning-based artificial PF to generate the optimal sequence for the reinforcement learning-based predictive control, achieving the real-time application, but it concentrated on the unstructured roads where the road PF was not utilized.}

{Overall, most of the aforementioned related work used the road PF and assigned a large potential value, making it unbreakable. Those studies generally assume that collision avoidance can be accomplished within the confine of the road, ignoring the ``blind alley'' problem. However, it might be necessary for a vehicle to maneuver in a way that makes it partially or entirely off the road to ensure safety in some cases, such as impending collision for a greater safety margin, avoiding secondary collisions, and unexpected intention of obstacles. Therefore, we focus on studying emergency collision avoidance that requires driving off the road to complete safe maneuvers.}


\section{Adaptive Potential Field\label{apf}}

In this section, we present the concept of the adaptive potential field, a paradigm that encompasses the road PF, obstacle PF, and target PF, collectively revolutionizing collision avoidance strategies for a safer driving experience.

\subsection{Road Potential Field}

The road PF usually prevents the vehicle from driving out of the road, and the vehicle can stay in the middle of the lane, ensuring safety. Therefore, we must establish the repulsive potential of the road edges and lane divider to enhance vehicle guidance and prevent unintended deviations. We use the following functions to represent the lane {divider} PF, $U_{lane}$, and road {edge} PF, $U_{{edge}}$, \cite{Wolf2008-ye}:
\begin{align}
\label{road_pf}
    U_{lane}&=
    \sum_{i=1}^{N_{lane}} A_{lane}\exp\left (-{{(Y-Y_{c,i})^{2}}\over{2\zeta^{2}}}\right),\\
    U_{{edge}}&=
    \frac{1}{2}\eta\left ({1\over Y-Y_{l,u}}\right)^{2},
\end{align}
where $N_{lane}$ determines the number of lanes, $A_{lane}$ denotes the maximum amplitude of the lane {divider} PF, $Y$ is the lateral position of the ego vehicle, $Y_{c,i}$ is the lateral position of the $i^{th}$ lane divider, $\zeta$ represents the slope of lane PF {that is directly proportional to the lane width and determines the rate at which the potential rises or falls.}, $\eta$ is a scale factor {that determines the steepness of the road PF}, and $Y_{l,u}$ denotes the lateral position of the lower and upper boundaries of the road. {It should note that the value of $A_{lane}$ is small when lane-change behavior is encouraged.}

\subsection{Obstacle Potential Field}

The obstacle PF is responsible for maintaining the ego vehicle at a safe distance from the obstacle. Therefore, the ego vehicle can be guided to perform lane-changing and/or braking maneuvers, based on the variation of the obstacle PF. The foundation of the obstacle PF, $U_{obs}$, is the probability density function (PDF) of the Gaussian distribution \cite{Lu2020-fm}:
\begin{equation} 
    U_{obs} =  
    \frac {{{A_{obs}}}}{{2\pi \sqrt {\left |{ \Sigma }\right |} }}{e^{\left ({{ - \frac {1}{2}\left ({{\ell - \mu } }\right)^{T}{\Sigma ^{ - 1}}{{\left ({{\ell - \mu } }\right)}}} }\right)}},
    \label{adaptive_pf}
\end{equation}
where $A_{obs}$ is the maximum amplitude of the obstacle PF, $\ell{=(s,\;d)^T}$ is the position of the ego vehicle in the Frenet frame \cite{Werling2010-bl} {that $s$ and $d$ are the tangential and normal directions in the Frenet coordinate}, and $\Sigma$ and $\mu$ are the covariance matrix and mean of the PDF, respectively. Therefore, we use Eq. (\ref{adaptive_pf}) to adapt the motion of the obstacle along with the longitudinal and lateral directions $U_{Aobs}$ as follows:
\begin{equation}
\label{obs_pf}
    U_{Aobs} = \sum_{{j=1}}^{N_{obs}}
    w_{1}U_{obs,{j}}^{1} +w_{2}U_{obs,{j}}^{2},
\end{equation}
where
    \begin{align*} w_{1}\in&[{0.5,1}],\qquad w_{2}=1-w_{1},\\ 
    \mu _{1}=&\left ({{s_{1},{d_{1}}} }\right)^{T},\quad \Sigma _{1} = \left [{ {\begin{array}{cccccccccccccccccccc} {{\sigma _{s_{1}}^{2}}}&0\\ 0&{{\sigma _{d_{1}}^{2}}} \end{array}} }\right],\\ 
    \mu _{2}=&\left ({{s_{2},{d_{2}}} }\right)^{T},\quad \Sigma _{2} = \left [{ {\begin{array}{cc} {{\sigma _{s_{2}}^{2}}}&0\\ 0&{{\sigma _{d_{2}}^{2}}} \end{array}} }\right],
\end{align*}
where {$N_{obs}$ denotes the number of obstacle vehicles,} $w_{1,2}$ is the weight factor, $\mu_1$ and $\mu_2$ denote the mean of the corresponding PDF that is related to obtaining the safe distance from the position of the obstacle in the Frenet frame, {$\Sigma_1$ and $\Sigma_2$ represent the covariance matrix of the corresponding PDF,} and $\sigma_{s,d}$ denotes the variance terms of the corresponding covariance matrix, which is combined with the longitudinal and lateral accelerations of the obstacle. The obstacle PF is shown in Fig. \ref{emergent_cutin}.

\subsection{Target Potential field}

The target PF constantly produces an attractive force to drive the ego vehicle forward or towards the target lane or position. Therefore, we simply used the following function to design the target PF, $U_{tar}$:
\begin{equation}
\label{tar_pf}
    U_{tar}= \frac{|X-X_{r}|+\left|Y-Y_{r}\right|}{\text{100}}
\end{equation}
where $(X_r,Y_r)$ denotes the position of the target location and $X$ is the longitudinal position of the ego vehicle. {It should be noted that Eq. (\ref{tar_pf}) is used in conventional driving scenarios where the emergency has not been sensed while driving toward the target position.} Thus, we can calculate the total PF by summing Eq. (\ref{road_pf})--(\ref{tar_pf}):
\begin{equation}
    U_{tol}=
    U_{lane}+U_{{edge}}+U_{Aobs}+U_{tar},
\end{equation}
We then applied the gradient descent method to obtain the desired path information:
\begin{equation}
    F_{tol}=-\nabla U_{tol}=-\begin{bmatrix}\frac{\partial U_{tol}}{\partial X}  &  \frac{\partial U_{tol}}{\partial Y}\end{bmatrix}^T,
\end{equation}
\begin{equation}
    \psi_{ref}=tan^{-1} \frac{F_{tol}(Y)}{F_{tol}(X)}.
\end{equation}
where $\psi_{ref}$ is the desired yaw angle. The overall APF is illustrated in Fig. \ref{emergent_cutin}, including the road structure and the obstacle.

\section{Clothoid Curve-based {Emergency-Stopping Path Planning}}\label{cceer}

In this section, we present the detailed design process of the proposed {ESPP}-based on the clothoid curve.

\subsection{Blind Alley Problem}
\begin{figure*}[t]
    \centering
    \includegraphics[width=\hsize]{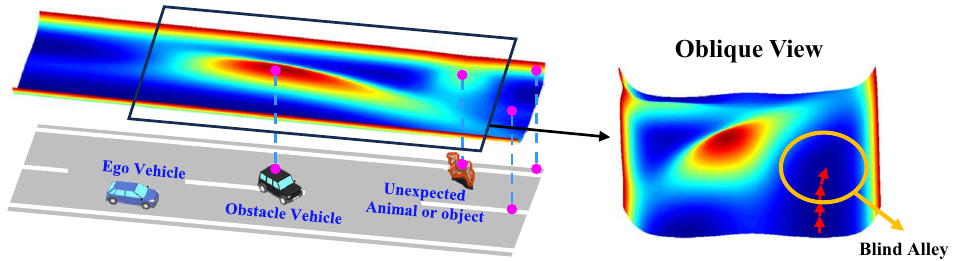}
    \caption{Emergency merging behavior of the adjacent vehicle (black) owing to the unexpected animal (brown): From the oblique view, the obstacle PF intersects with lower road PF, leading to a blind alley (local minima)}
    \label{emergent_cutin}
\end{figure*}
\begin{figure}
    \centering
    \includegraphics[width=\hsize]{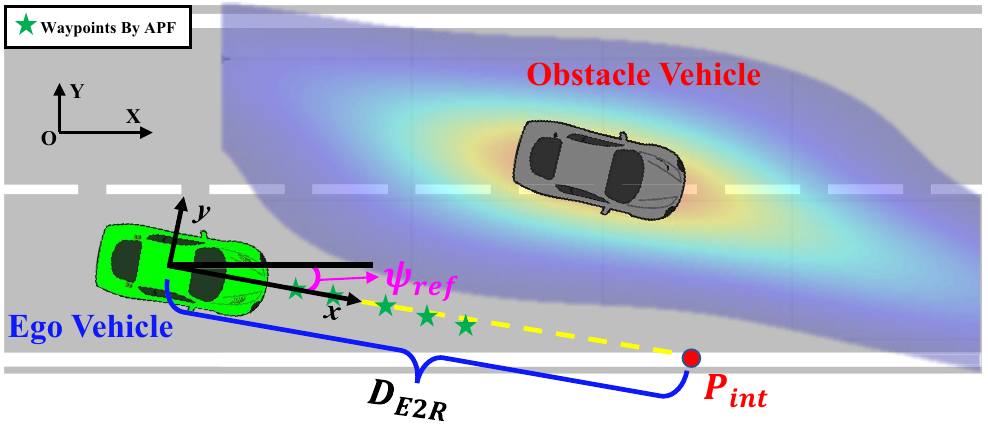}
    \caption{{Explanation of the emergency triggering estimation: With $\psi_{ref}$ given by APF, making an extension line from the current position of ego-vehicle (green) to intersect with the lower road boundary $P_{int}$; $D_{E2R}$ is the distance between green vehicle and $P_{int}$}}
    \label{emer_tri}
\end{figure}

Although the APF from Sec. III can model the collision risk with high accuracy by considering the longitudinal and lateral accelerations of the obstacle, it ignores the local minima problem with the assumption that the motion of the obstacle PF is known. However, traffic emergencies mostly occur because of the unpredictable motions of the surrounding obstacles, such as sudden deceleration and reckless lane changes \cite{Parker1995-ri}. As shown in \ref{emergent_cutin}, the sudden lane-changing obstacle with a full braking maneuver will lead to a rapidly expanding obstacle PF, which intersects with the road PF. Subsequently, the ``blind alley’’ phenomenon occurs because there are no feasible paths for the ego vehicle to track. This is a case of local minima. As described by Koren and Borenstein \cite{Koren1991-ar}, the local minima problem causes a trap situation for a mobile robot. However, considering the dynamic characteristics of the ego vehicle, it is difficult to stop at the local minima region owing to the large inertia. Instead, the ego vehicle could have driven into either the obstacle PF or road PF and received an excessive repulsive force \cite{Lin2022-op}, which can cause severe yawing and even vehicle crashes. {A simple way is to directly remove the road PF (the side to which the vehicle is heading) so that the ego vehicle can drive out of the road to obtain sufficient space for emergency braking and obstacle avoidance. However, the ego vehicle is prone to wheel slipping during emergency obstacle avoidance after removing the road PF due to the instantaneous disappearance of the road repulsive force. There has an imbalance in the virtual forces, leading to oversized repulsion from the obstacle PF experienced by the vehicle. In severe cases, they could lose control of the vehicle body. Although removing the road PF can eliminate the ``blind alley'' problem, it will lead to an imbalance of virtual forces because the repulsive force of the road PF disappears instantly, and the excessive repulsive force of the obstacle PF causes slipping, as shown in Fig. \ref{wheel_speed_withoutpf}. Therefore, we proposed a triggering estimation to detect the ``blind alley'' problem and then generate an emergency-stopping path based on the clothoid curve for completing a safe stop.
}

\subsection{Emergency Triggering Estimation}

{In this study, we propose an emergency triggering estimation to detect the blind alley problem that is described above.} As described in Section \ref{apf}, the APF can obtain $\psi_{ref}$ at each time step, which we can use to estimate the emergency. {Firstly, we use $\psi_{ref}$ to generate the local reference waypoints $(X_{ref},\;Y_{ref})$ for the controller to track, which is computed with a given step length $L$:}
\begin{equation}
    \begin{cases}
        X_{ref}=X+L\cos{\psi_{ref}}\\
        Y_{ref}=X+L\sin{\psi_{ref}}
    \end{cases}
    \label{waypoint}
\end{equation}
{Note that Eq. (\ref{waypoint}) is iterated in a control loop according to the number of required waypoints, and $L$ is usually dependent on the current speed and sampling time, for example, $L=VT_s$ where $V$ is the longitudinal speed of the ego vehicle and $T_s$ is the sampling time. Then, as depicted in Fig. \ref{emer_tri}, the green star-shape waypoints are produced by APF, which we can monitor the last waypoint $(X_{ref}^{last},\;Y_{ref}^{last})$, whether it reaches or cross the road edge at each time step. If yes,} we can make an extension line along the $\psi_{ref}$ (if $\psi_{ref} \neq 0$) angle from the current position of the ego-vehicle, as stated in Fig. \ref{emer_tri}. And the extension line will have an intersect point (denoted as $P_{int}$) with the lower road boundary. {Second}, we can measure the distance $D_{E2R}$ from the current position of the ego-vehicle to the intersect point. We mark the minimum braking distance of the ego-vehicle from the current position as $D_{brake}$. {It should note that $D_{brake}$ depends on multiple factors, including the reaction time, braking deceleration, road conditions, etc. D. Lyubenov \cite{Lyubenov2011-tv} has summarized the empirical formula to compute the minimum braking distance in accident investigation the cases for emergency braking behavior.} {If $D_{E2R}$ is greater than or equal to $D_{brake}$, the ego vehicle can promptly stop before reaching the road PF. In this case, the ESPP computation will not be triggered, and the ego vehicle can regain its orientation to follow the obstacle vehicle if the obstacle vehicle returns to regular driving after an emergency lane change.} Otherwise, the {ESPP} computation will be triggered to navigate the ego-vehicle to a safe stop. {In general, the overall process of the emergency triggering estimation is summarized in Fig. \ref{flow_chart}.}
\begin{figure}
    \centering
    \includegraphics[width=0.8\hsize]{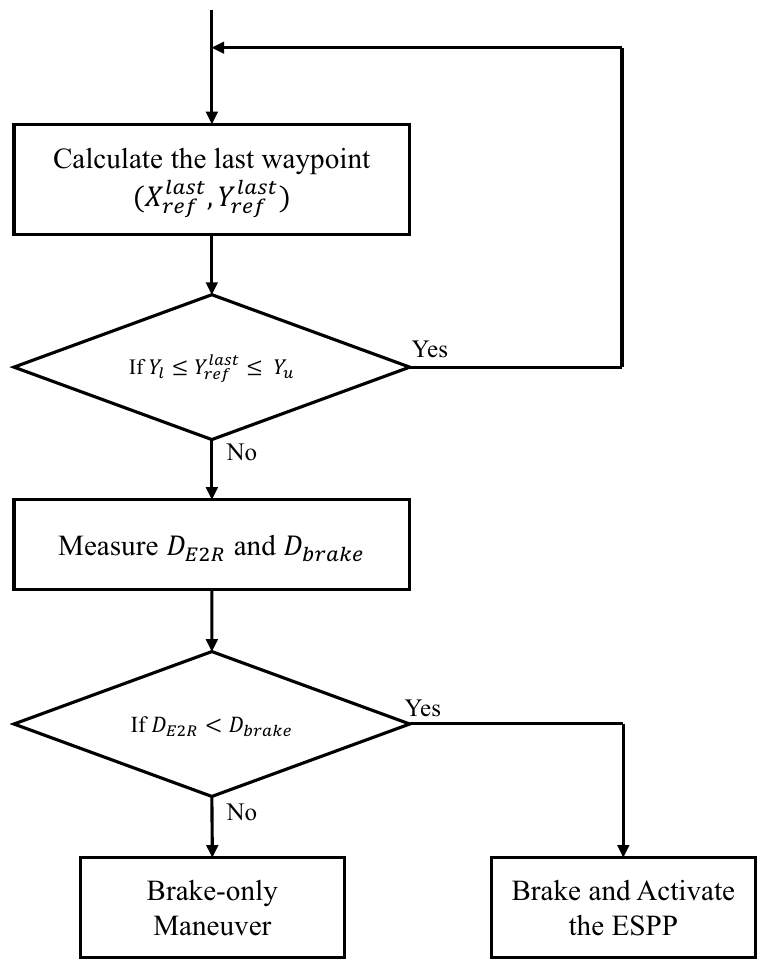}
    \caption{{System flow chart of the proposed emergency triggering estimation}}
    \label{flow_chart}
\end{figure}

\subsection{Clothoid Curve}
\begin{figure*}[t]
    \centering
    \includegraphics[width=\hsize]{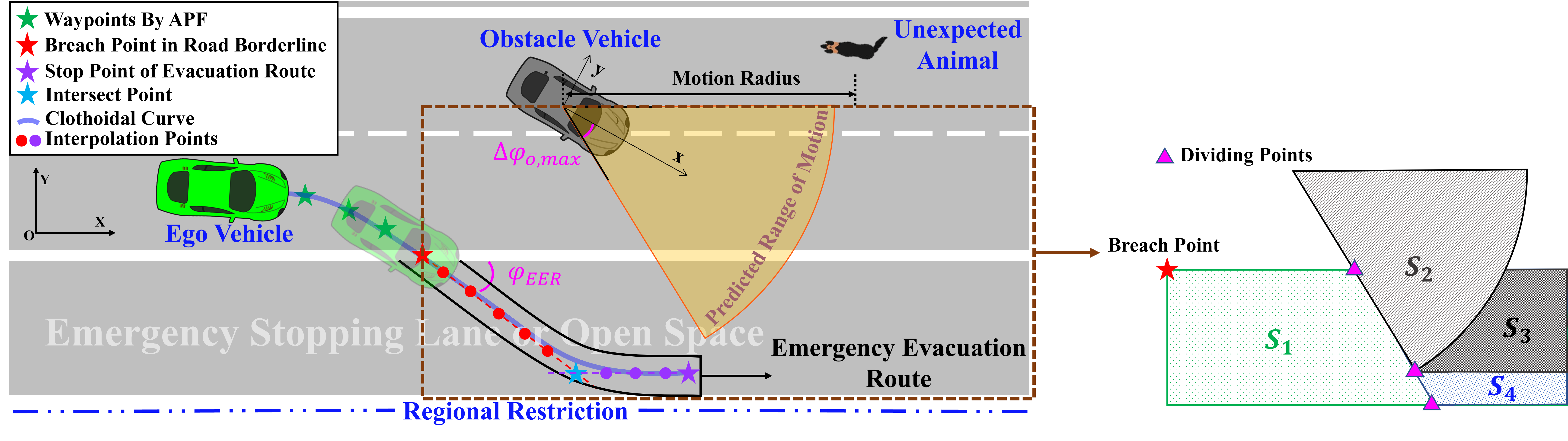}
    \caption{{Emergency-Stopping Path Planning on the expressway}: Regionalization within the brown dashed box ($S_1,\; S_2,\; S_3,\; S_4$); three star-shape points are then obtained (red, blue, purple); interpolation points (circle) are used for computing the clothoid curve (light blue).}
    \label{route_explain}
\end{figure*}

{To tackle the "blind alley" situation, we propose an {ESPP method} based on the clothoid curve that is also known as cubic (3rd order) polynomial.} The clothoid curve is usually used in waypoint tracking and highway road design to generate an easy-to-follow path with linear curvature \cite{Cheng2011-xe, Ozguner2011-ik}. The typical representation of the clothoid curve is denoted as \cite{Jeon2015-vd}:
\begin{equation}  
    f_{c}({r})=c_{0}+c_{1}{r}+c_{2}{r}^{2}+c_{3}{r}^{3}, 
\end{equation}
where
\begin{equation*}
    \kappa=6c_{3}s+2c_{2},
\end{equation*}
where ${r}$ denotes the arc length, $c_0$ denotes the lateral offset from the lane center at ${r}=0$, $c_1$ denotes the heading angle error at ${r}=0$, $\kappa$ denotes the curvature of the clothoid curve, $2c_2$ denotes the road curvature at ${r}=0$, and $3c_3$ denotes the curvature rate. To estimate the clothoidal coefficients, at least 4 waypoints are required for curve fitting, as illustrated below:
\begin{equation} 
    \begin{bmatrix}
        1 &x_{1} &{x_{1}}^{2} &{x_{1}}^{3}\\ 1 &x_{2} &{x_{2}}^{2} &{x_{2}}^{3}\\ \vdots &\vdots &\vdots &\\ 1 &x_{N} &{x_{N}}^{2} &{x_{N}}^{3}\end{bmatrix}
        \begin{bmatrix}c_{0}\\ c_{1}\\ c_{2}\\ c_{3}\end{bmatrix}=\begin{bmatrix}f(x_{1})\\ f(x_{2})\\ \vdots\\ f(x_{N})\end{bmatrix}\approx \begin{bmatrix}y_{1}\\ y_{2}\\ \vdots\\ y_{N}
    \end{bmatrix}.
    \label{cf}
\end{equation}
For $N\geq4$ waypoints, we calculate the coefficients, $\mathbf{C}=\begin{bmatrix}c_0& c_1& c_2& c_3\end{bmatrix}^T$. $(x_N,y_N)$ is defined with reference to the vehicle coordinates ${xyz}$, whereas the waypoints from the APF and GPS sensors are defined with reference to the local projected coordinates ${XYZ}$. Therefore, the coordinate transformation is required to present the waypoints in the vehicle coordinates for curve fitting, given by:
\begin{equation}
    \begin{bmatrix}x_{i}\\ y_{i}\\ 1\end{bmatrix}=
    \overbrace{\begin{bmatrix}\cos{\phi} &-\sin{\phi} &0\\ \sin{\phi} &\cos{\phi} &0\\ 0 &0 &1\end{bmatrix}}^{\text{Rot}(z,{\phi})} \overbrace{\begin{bmatrix}1 &0 &d_{x}\\ 0 &1 &d_{y}\\ 0 &0 &1\end{bmatrix}}^{\text{Trans}(d_{x}, d_{y})}\begin{bmatrix}X_{i}\\ Y_{i}\\ 1\end{bmatrix},
\end{equation}
for $i=1,2,\ldots,N$, where
\begin{equation*}
    {\phi}=-\psi, \quad d_{x}=-{X}, \quad d_{y}=-{Y}.
\end{equation*}
{with $\psi$ is the heading angle of the ego vehicle.} Typically, the clothoid curve is used to fit waypoints from the leading vehicle in a car-following scenario. However, in this study, we used a clothoid curve to fit a set of specific waypoints to create the {ESPP}. 

\subsection{{Emergency-Stopping Path Planning}}
\begin{algorithm}[t]
	\renewcommand{\algorithmicrequire}{\textbf{Input:}}
	\renewcommand{\algorithmicensure}{\textbf{Output:}}
	\caption{Selecting Stop Point of the {ESPP}}
	\label{alg1}
	\begin{algorithmic}[1]
		\STATE Initialization:$ P_{bp}, P_{sp}, D_{manha} \leftarrow \mathbf{0}$
		\STATE Compute $P_{sp}$ $\leftarrow$ $\{x_{sp},y_{sp}\}$
		\STATE Compute $S_2$ based on Eq. (\ref{s2})
        \FOR{$x_{sp,cand} \in [x_{bp},x_{bp}+D_{brake}]$}
            \FOR{$y_{sp,cand} \in [Y_l,Y_{esl}-l_w/2]$}
                \IF{$\psi_{o}\leq 0$}
            		\STATE Obtain $P_{dp}^i$ $\leftarrow$ $\{x_{dp}^i,y_{dp}^i\}$
            		\STATE Optimize $D_{manha}$ Eq. (\ref{optimization}) for $P_{sp}^*$
            	\ELSE
            		\STATE Optimize $D_{manha}$ Eq. (\ref{optimization}), without Eq. (13a) and (13b) for  $P_{sp}^*$,
            	\ENDIF
            \ENDFOR
        \ENDFOR
		\ENSURE  selected stop point ${P_{sp}^*}$
	\end{algorithmic}  
\end{algorithm}

As depicted in Fig. \ref{route_explain}, when the adjacent vehicle suddenly merges without a pre-warning, the waypoints produced by the APF from the ego vehicle will lead to the lower road borderline. In this situation, the ego vehicle hits the PF of the lower road boundary, even with complete braking. An excessive repulsive force is imposed on the ego vehicle because the road PF restricts the vehicles from driving out of the road, leading to a severe heading oscillation of the ego vehicle. Herein, opening a breach from the road PF is necessary to navigate the ego vehicle to stop at a non-conflicting position. From the waypoints (green stars) generated by the APF, we can estimate the heading angle of the ego vehicle when it drives toward the road boundary. The estimated heading angle is then used to find the waypoint (red star) intersecting the road boundary as the breach spot. Subsequently, the most crucial step is to determine the stop point (purple star). Thus far, the selection of the stop point should consider the following constraints:
\begin{itemize}
    \setlength{\itemsep}{0pt}
    \setlength{\parsep}{0pt}
    \setlength{\parskip}{0pt}
    \item The stop point should not be within the predicted motion range of the obstacle.
    \item The stop point should not be in the restricted region (denoted as the blue double-dotted line).
    \item The total length of the {ESPP} should consider the distance and time from braking to a safe stop.
\end{itemize}
Based on these requirements, we can regionalize the selected area (denoted by the brown dashed frame in Fig. \ref{route_explain}) by two-dimensional geometry. The predicted motion range of the obstacle vehicle is denoted by $S_2$ as follows:
\begin{equation}
    S_2=\frac{1}{2}\alpha R^2,
    \label{s2}
\end{equation}
where
\begin{equation*}
    \alpha=2\Delta\psi_{o,max},\quad R=N_{p} V_{obs} T_s,
\end{equation*}
where {$\alpha$ is the central angle and $R$ is the motion radius of the obstacle vehicle.} $\Delta\psi_{o, max}$ denotes the maximum heading angle of the {obstacle} vehicle {and $V_{obs}$ is the velocity of the obstacle vehicle.} $N_p$ denotes the prediction horizon {(consistent with the MPC controller)}, and $T_s$ denotes the sampling time. {In this study, we use approximate estimation to predict the motion range of the obstacle vehicle. At each time step, the ego vehicle measures $\Delta\psi_{o, max}$ and $V_{obs}$ that we can roughly estimate the possible motion range considering the maximum mechanical steering limitation. Then,} we can then calculate the coordinates of the three dividing points (denoted as $P_{dp}^1$, $P_{dp}^2$, and $P_{dp}^3$, from top to bottom) using the geometric solution. Therefore, we should assign the stop point (denoted as $P_{sp}$) outside $S_2$ and $S_3$, because the collision risk is higher than $S_1$ and $S_4$. Moreover, if $P_{sp}$ is located in $S_3$, it leads to a tortuous curve. Therefore, we hope that $P_{sp}$ can be selected from $S_1$, which is the safest area. Nevertheless, if the length of the {ESPP} is less than the minimum braking distance of the ego vehicle, $S_4$ should be considered for $P_{sp}$. Finally, we can arrange it into an optimization problem with the following constraints:
\begin{alignat}{2}
    \centering
    \label{optimization}
    \max_{x_{sp},y_{sp}} &|x_{sp}-x_{bp}|+|y_{sp}-y_{bp}|\\
    \label{constraints}
    \mathrm{s.t.} 
    &-\psi_{{ESPP}}< \arctan(\frac{y_{sp}-y_{bp}}{x_{sp}-x_{bp}})< 0 \quad ,& \tag{14a}\\
    &
    \begin{cases}
        P_{sp}=\{x_{sp},y_{sp}\}\in S_1, & {if \quad {L_{ESPP}}\geq D_{brake}},\\
        P_{sp}=\{x_{sp},y_{sp}\}\in S_4, & {if \quad {L_{ESPP}}<D_{brake}},
    \end{cases}& \tag{14b}
\end{alignat}
where $\psi_{{ESPP}}$ denotes the heading angle of the {ESPP}, ${L_{ESPP}}$ denotes the length of the {ESPP}, and $D_{brake}$ denotes the minimum braking distance. {The objective of Eq. (\ref{optimization}) is to maximize the Manhattan distance between $P_{sp}$ and the breach point, $P_{bp}$, because we need to stop as far as possible, considering that the car needs a sufficient distance to brake and also needs to be as far away from dangerous areas as possible. Therefore, we have specified the safe zone ($S_1$ and $S_4$) in the constraints (14a) and (14b) with regionalization. By solving the above constrained objective function, we can get the optimal point for $P_{sp}$.} The detailed procedure for selecting $P_{sp}$ is presented in Algorithm \ref{alg1}. $Y_{esl}$ refers to the lower boundary of the emergency stopping lane, and $l_w$ is the vehicle's width. Provided that if we determine $P_{bp}$ and $P_{sp}$, we can compute an intersection point, $P_{ip}$, using two line segments extending from points $P_{bp}$ and $P_{sp}$. The tilt angle of the line segment extending from $P_{bp}$ is equal to the heading angle of the ego vehicle. By doing this, we can ensure that the ego vehicle will not have serious fluctuations in heading when entering the {ESPP}. Moreover, the line segment from $P_{sp}$ should be parallel to the road direction, which conforms to traffic rules when using the emergency stopping lane (ESL). Therefore, we can interpolate certain waypoints (red and purple dots) between the two line segments for curve fitting. We subsequently re-formulate Eq. (\ref{cf}) through the waypoints belonging to three categories: APF, the first line segment (red dotted line), and the second line segment (purple dotted line).
\begin{equation} 
    \displaystyle \underbrace{\begin{bmatrix} 1 & x_{1,apf} & x_{{1,apf}}^2 & x_{{1,apf}}^3\\ 1 & x_{2,apf} & x_{2,apf}^2 & x_{2,apf}^3\\ 
    \vdots & \vdots & \vdots & \vdots\\
    1 & x_{{M}-1,bp} & x_{{M}-1,bp}^2 & x_{{M}-1,bp}^3\\
    1 & x_{{M},bp} & x_{{M},bp}^2 & x_{{M},bp}^3\\
    \vdots & \vdots & \vdots & \vdots \\
    1 & x_{{N-2},ip} & x_{{N-2},ip}^2 & x_{{N-2},ip}^3\\
    1 & x_{{N-1},ip} & x_{{N-1},ip}^2 & x_{{N-1},ip}^3\\
    1 & x_{sp} & x_{sp}^{2} & x_{sp}^{3} \end{bmatrix}}_{\mathbf{X}}
    \underbrace{\begin{bmatrix} 
    c_{0}\\ c_{1}\\ c_{2}\\ c_{3} \end{bmatrix}}_{\mathbf{C}_{v}}=
    \underbrace{\begin{bmatrix} 
    f(x_{1,apf})\\ f(x_{2,apf})\\ \vdots\\ f(x_{{M}-1,sp})\\ f(x_{{M},sp})\\ \vdots\\ f(x_{{N-2},ip})\\ f(x_{{N-1},ip})\\ fx_{sp}) 
    \end{bmatrix}}_{\mathbf{F}}
    \label{hybr_cf}
\end{equation}
\begin{algorithm}[t]
	\renewcommand{\algorithmicrequire}{\textbf{Input:}}
	\renewcommand{\algorithmicensure}{\textbf{Output:}}
	\caption{Curve Fitting with Hybrid Waypoints}
	\label{algo2}
	\begin{algorithmic}[1]
		\STATE Initialization:$P_{ip}, P_{new} \leftarrow \mathbf{0}$ and $N_{b2i}, N_{i2s}, p_{num}$
		\STATE Compute $P_{ip}$ $\leftarrow$ $Intersect(P_{bp},P_{sp})$
		\STATE Interpolate $N_{b2i}$ Waypoints between $P_{bp}$ and $P_{ip}$
		\STATE Interpolate $N_{i2s}$ Waypoints between $P_{ip}$ and $P_{sp}$
		\STATE Conduct Curve Fitting based on Eq. (14) and (15), respectively.
		\STATE Obtain $\mathbf{c}^*$ based on Eq. (16) and (17)
		\WHILE{$p_{num}$}
		    \STATE Compute $U_{{ESPP}}$ based on Eq. (18), (19), and (20):
		    \STATE Obtain $P_{new}$ by Gradient Descent 
		    \STATE $p_{num} \leftarrow p_{num}-1$
		\ENDWHILE
		\ENSURE  new waypoints ${P_{new}}$
	\end{algorithmic}  
\end{algorithm}

\noindent {where $P_{apf}=(x_{apf},\;f(x_{apf}))$ is the position pair of the waypoints from the APF, $P_{bp}=(x_{bp},\;f(x_{bp}))$ denotes the position pair of the breach points, $P_{ip}=(x_{ip},\;f(x_{ip}))$ represents the position pair of the intersection points, and $P_{sp}=(x_{sp},\;f(x_{sp}))$ is the position pair of the stop point.} Subsequently, we can obtain the clothoidal coefficients through the method of least-squares. Considering that the least square method cannot make the curve pass through all the selected waypoints accurately, it might lead to low accuracy at the end of the curve. Therefore, we introduce a weight matrix $W$ into Eq. (\ref{hybr_cf}), as follows:
\begin{equation}
    W\mathbf{X}\mathbf{C}_{v}=W\mathbf{F},
    \label{simple_cf}
\end{equation}
where
\begin{equation*} 
    W=diag{(w_{1}, w_{2},\ \ldots, w_{N})}_{N\times N}.
\end{equation*}
Thus, the fitting accuracy at the end of the curve can be improved by adjusting the weight matrix to stop the ego vehicle at the desired location. Subsequently, the clothoidal coefficients are obtained as follows:
\begin{equation}
    \mathbf{C}_{v}=((W\mathbf{X})^{T}W\mathbf{X})^{-1}(W\mathbf{X})^{T} W\mathbf{F}.
    \label{coe}
\end{equation}
Eq. (\ref{coe}) still does not consider vehicle dynamics; thus, it cannot ensure that the path is always trackable for the ego vehicle, particularly with different dynamic characteristics \cite{Kang2017-jk}. To overcome this, we reformulate Eq. (\ref{coe}) to a standard QP form with the following constraints:
\begin{alignat}{2}
    \label{qp_cost}
    \textbf{c}^* \, = \, &
        \arg\min_{\textbf{c}\in C}\frac{1}{2}\textbf{c}^T\mathbf{H}\textbf{c}+\mathbf{f}^T\textbf{c}\\
    & \mathrm{s.t.} \quad \textbf{c}_{min}\preceq \textbf{c}\preceq \textbf{c}_{max},\tag{18a}
\end{alignat}
where
\begin{align*}
    &H=\textbf{X}^TW\textbf{X},\,\,f=-\textbf{X}^TW\textbf{F},\\
    &\textbf{c}_{min}=\begin{bmatrix} e_{y}^{min}& e_{\psi}^{min}& -\frac{\omega_{max}^2}{2\upsilon g}& -\frac{\dot{\kappa}_{max}}{6} \end{bmatrix}^T,\\
    &\textbf{c}_{max}=\begin{bmatrix} e_{y}^{max}& e_{\psi}^{max}& \frac{\omega_{max}^2}{2\upsilon g}& \frac{\dot{\kappa}_{max}}{6} \end{bmatrix}^T,\,\,\kappa_{max}=1/R_{min},
\end{align*}
\noindent where $e_{y}^{min,max}$ denotes the minimum and maximum lateral position errors, and $e_{\psi}^{min,max}$ represent the minimum and maximum yaw angle errors, respectively; $\omega_{max}$ is the maximum angular velocity of the ego vehicle, $\upsilon$ is the friction coefficient, $g$ is the gravitational acceleration, $\kappa_{max}$ denotes the maximum curvature, and $\dot{\kappa}_{max}$ determines the maximum curvature rate to empirically indicate that the ego vehicle can only steer the wheel within a limited range under its current speed \cite{Scheuer1997-bv}, {$R_{min}$ is the minimum turning radius that is determined by the vehicle model.}  {The objective of Eq. (\ref{qp_cost}) is to compute the optimal clothoidal coefficients under specific constraints. It should be noted that Eq. (17) only considers the fitting accuracy of the waypoints. However, the obtained waypoints are not guaranteed to be trackable. Furthermore, the numerical values of the clothoidal coefficients vary around zero (e.g., $c_3$), which can easily affect the shape of the curve. Therefore, we solve the Eq. (\ref{qp_cost}) with constraints (18a) to consider the fitting errors, angular speed, and curvature rate that ensures the clothoid curve is properly generated and conforms to vehicle dynamics at the current speed.}

Further, we can use the obtained clothoid curve to establish the PF for the {ESPP} through the following formulas, including the lower and upper boundary PFs, ($U_{{ESPP}}^{lb}$ and $U_{{ESPP}}^{rb}$), and the attractive PF, $U_{{ESPP}}^{attr}$, as follows:
\begin{equation}
    \begin{split}
        & U_{{ESPP}}^{lb}=\\
        & A_{e}\left(1-e^{-b_w sign(y-f_{cr}(x))\sqrt{\left(\frac{y-b_{y}}{{m_y}}-x\right)^{2}+(f_{cr}(x)-y)^{2}}}\right)^{2}
    \end{split}
    \label{new_rpf_ub}
\end{equation}
\begin{equation}
    \begin{split}
        & U_{{ESPP}}^{rb}=\\
        & A_{e}\left(1-e^{b_w sign(y-f_{cl}(x))\sqrt{\left(\frac{y-b_{y}}{{m_y}}-x\right)^{2}+(f_{cl}(x)-y)^{2}}}\right)^{2}
    \end{split}
    \label{new_rpf_rb}
\end{equation}
\begin{equation}
    U_{{ESPP}}^{attr}=\frac{1}{2}\xi D(X_d,Y_d)^2,\\
    \label{new_attr}
\end{equation}
\begin{figure}[t]
    \centering
    \includegraphics[width=\hsize]{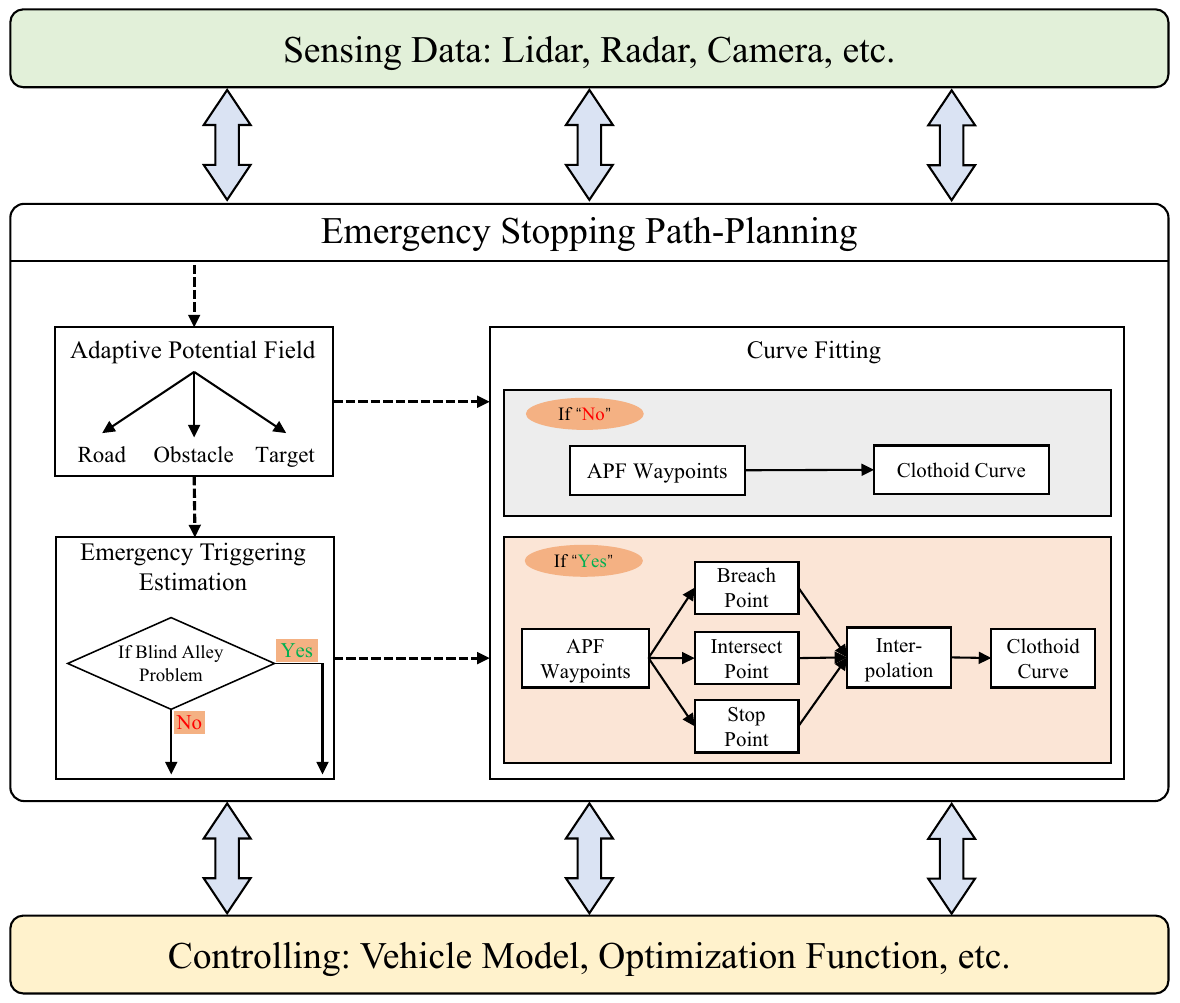}
    \caption{{Detailed system architecture of the proposed ESPP method}}
    \label{sys_archi}
\end{figure}
where 
\begin{align*} 
    b_y=f_{cr}(x)-{m_y} x,\quad {m_y}=-\frac{1}{\dot{f}_c(x)},
\end{align*}
with $A_{e}$ is the maximum amplitude of the {ESPP}'s PF, {$(x,\;y)$ is the position pair of the ego vehicle in vehicle coordinate,} $b_w$ denotes the parameter controlling the road PF width \cite{Snapper2018-mn}, {${f}_c(x)$ is the optimal curve that is obtained from Eq. (\ref{qp_cost}), $f_{cr}(x)$ and $f_{cl}(x)$ are the right and left boundaries of the {ESPP}, respectively, which are acquired by shifting ${f}_c(x)$}, $\xi$ is the influence factor of the attractive PF, $U_{{ESPP}}^{attr}$; and $D(X_d, Y_d)$ denotes the Euclidean distance between the ego vehicle and the temporary target point from the clothoid curve. {Note that Eqs. (\ref{new_rpf_ub}) and (\ref{new_rpf_rb}) are used to establish an impenetrable PF that keeps the vehicle tracking around the ESPP trajectory. In addition, we use Eq. (\ref{new_attr}) to produce the attractive force instead of Eq. (\ref{tar_pf}) when the ``blind alley'' problem is detected, considering Eq. (\ref{new_attr}) can generate a more potent force to lead the ego-vehicle to follow the optimized clothoid curve.} Finally, we can model the 3D PF of the {ESPP} and obtain the emergency collision-free path by repeating the process described in Section \ref{adaptive_pf}, as shown in Fig. \ref{3D_pf_eer}. We also present the curve-fitting process using hybrid points in Algorithm \ref{algo2}. Therefore, the {ESPP} can guide the ego vehicle to conduct emergency collision avoidance and safe stop maneuvers. {Overall, the proposed system architecture of the ESPP method is depicted in Fig. \ref{sys_archi}, which indicates the workflow of the internal modules.}
\begin{figure}[t]
    \centering
    \includegraphics[width=0.3\hsize]{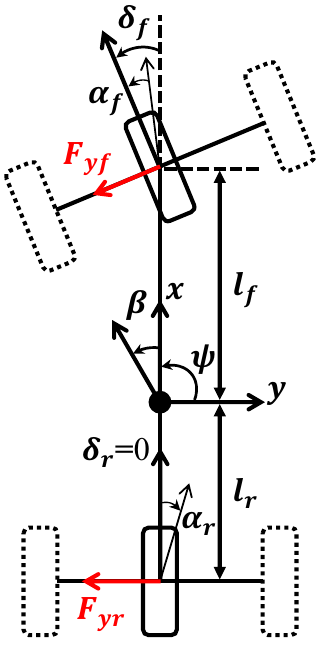}
    \caption{Vehicle lateral dynamics model for path tracking}
    \label{vehicle_model}
\end{figure}

\section{MODEL PREDICTIVE CONTROLLER}

In this section, we will comprehensively illustrate the specific MPC for path-tracking tasks, encompassing vehicle dynamics, thorough constraints analysis, detailed optimization function formulation, and effective quadratic programming techniques.

\subsection{Vehicle Dynamics Model}

One of the advantages of the MPC is that it has a built-in vehicle dynamics model, which can predict the motion states of the vehicle within a given prediction horizon. The four-wheel vehicle dynamics model is usually simplified as a bicycle model by assuming the front (rear) two wheels have the same steering. Another important postulation is that the bicycle model is two-degree-of-freedom (2DOF) and neglects the angular momentum of the vehicle body in roll, yaw, and pitch, as shown in Fig. \ref{vehicle_model}. The vehicle dynamics model can then be mathematically formulated by the following equations \cite{Rajamani2011-ua}:
\begin{align} 
    mV(\dot{\beta}+\dot{\psi})=&\,F_{yf}+F_{yr} \\ I_{z}\ddot{\psi}=&\,l_{f}F_{yf}-l_{r}F_{yr} 
\end{align}
where
\begin{align*} 
    \dot{\beta}=&\,\frac{-(C_{r}+C_{f})}{mV}\beta+\left(\frac{C_{r}l_{r}-C_{f}l_{f}}{mV^{2}}-\text{1}\right)\dot{\psi}+\frac{C_{f}}{mV}\delta_f,\\ \ddot{\psi}=&\,\frac{C_{r}l_{r}-C_{f}l_{f}}{I_{z}}\beta-\frac{C_{r}l_{r}^{2}+C_{f}l_{f}^{2}}{I_{z}V}\dot{\psi}+\frac{C_{f}l_{f}}{I_{z}}\delta_f,\\
    F_{yf}=&\,C_{f}\alpha_{f}=C_{f}\left(\delta_f-\beta-\frac{l_{f}\dot{\psi}}{V}\right),\\ F_{yr}=&\,C_{r}\alpha_{r}=C_{r}\left(-\beta+\frac{l_{r}\dot{\psi}}{V}\right) 
\end{align*}
with $m$ is the vehicle mass, $\beta$ and $\dot{\beta}$ denote the sideslip angle and the sideslip rate of the vehicle, respectively. $\dot{\psi}$ denotes the yaw rate of the vehicle, $I_z$ is the yaw moment of inertia, $l_f$ and $l_r$ are the front and rear distances from CG (center of gravity) to front and rear tires, respectively. $F_{yf}$ and $F_{yr}$ denote the front and rear lateral tire forces, respectively. $C_f$ and $C_r$ represent the cornering stiffness of the front and rear tires respectively. $\alpha_f$ and $\alpha_r$ are the tire slip angles. $\delta_f$ is the front tire steering angle of the vehicle. To facilitate matrix operation in MPC, we should transform the vehicle dynamics model into discrete state-space formulation \cite{Eckert2011-gt}, as shown below:
\begin{align}
    \boldsymbol{x}(k+1)=&\,\mathbf{A}\boldsymbol{x}(k)+\mathbf{B}\delta_f \\ 
    \boldsymbol{y}=&\,\mathbf{C}\boldsymbol{x}(k)
\end{align}
where
\begin{align*}
    \boldsymbol{x}=&\left[Y\quad\beta\quad\psi\quad\dot{\psi}\right]^{T},\\
    \mathbf{A}=&\left[\begin{array}{@{}cccc@{}} \text{1}&T_sV&T_sV&\text{0} \\ \text{0}&1-T_s\frac{C_{r}+C_{f}}{mV}&\text{0}&T_s\frac{C_{r}l_{r}-C_{f}l_{f}}{mV^{2}}-T_s \\ \text{0}&\text{0}&\text{1}&T_s \\ \text{0}&T_s\frac{C_{r}l_{r}-C_{f}l_{f}}{I_{z}}&\text{0}&1-T_s\frac{C_{r}l_{r}^{2}+C_{f}l_{f}^{2}}{I_{z}V}\end{array}\right] \\ \mathbf{B}=&\left[\begin{array}{cccc} \text{0} & T_s\frac{C_{f}}{mV} & \text{0} & T_s\frac{C_{f}l_{f}}{I_{z}}\end{array}\right]^T\!,\ \mathbf{C}=\left[\begin{array}{@{}cccc@{}} \text{1}&\text{0}&\text{0}&\text{0} \\ \text{0}&\text{1}&\text{0}&\text{0} \\ \text{0}&\text{0}&\text{0}&\text{1}\end{array}\right].
\end{align*}
Therefore, we use the above discrete state-space model for the motion prediction and optimization.

\subsection{Optimization Function}

The significance of the optimization cost function is to enable the controller to follow the desired trajectory quickly and smoothly, which minimizes the deviation of the system state variables and the control variables. Therefore, three items constitute the complete optimization function: (\romannumeral1) Deviation between state variables and desired trajectory; (\romannumeral2) Variation of the control variables; (\romannumeral3) Penalization on the violation of the constraints \cite{Falcone2007-gj}, as illustrated below:
\begin{align} 
    \boldsymbol{J} =
    &\sum\limits_{\boldsymbol{k} = 1}^{{\boldsymbol{N}_{\boldsymbol P}}} {\left\| {{\boldsymbol{y}(k|t)} - {\boldsymbol{y}^d(k|t)}} \right\|_{\boldsymbol Q}^2}\nonumber \\ &+ \sum\limits_{k = 0}^{{\boldsymbol{N}_{\boldsymbol C}} - 1} {\left\| {\Delta {\mathrm{u}(k|t)}} \right\|_{\boldsymbol R}^2} + \lambda \epsilon,
    \label{cost_func_inc}
\end{align}
where $N_c$ denotes the control horizon, $Q$ and $R$ are the weighting matrices, ${\boldsymbol{y}(k|t)}$ means the model outputs at $k$ steps from the current time $t$, ${\boldsymbol{y}^d(k|t)}$ denotes the desired reference signals at $k$ steps from the current time $t$, {$\Delta\mathrm{u}(k|t)$ is the increment of the $u(k|t)=\delta_f$.} $\lambda$ is a weight coefficient and $\epsilon$ is a slack variable. After that, we need to add conditional constraints to the optimization function to reflect the vehicle's mechanical structure limitations, road structure, etc. Thus, the following constraints are considered:
\begin{gather}
    \Delta \delta_f^{min}
    \leq \Delta \mathrm{u}(k|t) \leq \Delta \delta_f^{max},\\
    \delta_f^{min} 
    \leq \Delta \mathrm{u}(k|t) + \mathrm{u}(k-1|t) \leq  \delta_f^{max},\\ 
    \boldsymbol{y}_{min} \leq \boldsymbol{y}(k|t) \leq \boldsymbol{y}_{max},
\end{gather}
where $\Delta \delta_f^{min}$ and $\Delta \delta_f^{max}$ represent the minimum and maximum incremental front tire steering angles, respectively. $\delta_f^{min}$ and $\delta_f^{max}$ are the minimum and maximum front tire steering angles, respectively. $\boldsymbol{y}_{min}=\begin{bmatrix}
Y_{l} & \beta_{min} & \dot{\psi}_{min}
\end{bmatrix}^T$ and $\boldsymbol{y}_{max}=\begin{bmatrix}
Y_{u} & \beta_{max} & \dot{\psi}_{max}
\end{bmatrix}^T$ denote the minimum and maximum output constraints, respectively, {where $\beta_{min,\;max}$ is the minimum and maximum sideslip angle and $\dot{\psi}_{min,\;max}$ is the minimum and maximum yaw rate of the vehicle.} It should be noted that the position constraints in $\boldsymbol{y}_{min}$ and $\boldsymbol{y}_{max}$ only consider the regular road structure such as the lower and upper road boundaries. However, referring to the traffic emergency in Fig. \ref{emergent_cutin}, the usual road constraints in MPC will prevent the {ESPP} method from working; because the {ESPP} method is to open a breach point from the road boundary so that the vehicle can be navigated to stop at a safe region. Therefore, to guarantee the {ESPP} method still work in traffic emergencies, we make adjustments in the road structure constraints of the MPC controller accordingly. The overall optimization function with constraints is demonstrated as follows:
\begin{align} 
    \label{cost_func}
    & \mathop{{\rm{min}}}\limits_{\Delta \mathrm{u},\epsilon} \,\boldsymbol{J}(\boldsymbol{x}(k|t),\Delta \mathrm{u}(k|t))\qquad\\ {\rm{s}}{\rm{.t}}{\rm{.}}\,
    &\boldsymbol{x}(k+1)=\,\mathbf{A}\boldsymbol{x}(k)+\mathbf{B}\delta_f\tag{30a}\\ 
    &\boldsymbol{y}=\,\mathbf{C}\boldsymbol{x}(k)\tag{30b}\\ 
    &\Delta \delta_f^{min}
    \leq \Delta \mathrm{u}(k|t) \leq \Delta \delta_f^{max},\tag{30c}\\
    &\delta_f^{min} 
    \leq \mathrm{u}(k|t) \leq  \delta_f^{max},\tag{30d}\\ 
    &\mathrm{u}(k|t) = \mathrm{u}(k-1|t) + \Delta \mathrm{u}(k|t)\tag{30e}\\ 
    &k = t, \ldots,t + {N_p} - 1,\nonumber\qquad\qquad\\
    &\begin{cases}
        \boldsymbol{y}_{min} \leq \boldsymbol{y}(k|t) \leq \boldsymbol{y}_{max},& { D_{E2R}>D_{brake}}\\
        \boldsymbol{y}_{min}^{emer} \leq \boldsymbol{y}(k|t) \leq \boldsymbol{y}_{max},& { D_{E2R}\leq D_{brake}}
    \end{cases},\tag{30f}
\end{align}
where
\begin{equation*}
    \boldsymbol{y}_{min}^{emer}=
    \begin{bmatrix}
        Y_{esl} & \beta_{min} & \dot{\psi}_{min}
    \end{bmatrix}^T
\end{equation*}

Note that constraint (30f) is determined by the emergency triggering estimation introduced in Sec. \ref{cceer}-B. Furthermore, we can transform Eq. \ref{cost_func} into the standard quadratic programming (QP) form and then choose the preferred QP solver to reduce time consumption \cite{Gao2014-bg}.

\section{SIMULATION RESULTS\label{results}}

This section introduces the initial environmental setup and detailed simulation results. We conducted a comprehensive co-simulation study using MATLAB/Simulink and CarSim simulator to verify the proposed {ESPP}-based APF method. To compare the effectiveness of the proposed method, we computed four types of path planners as follows: (i) conventional PF with constant speed (denoted as CPF with CS) \cite{Lin2022-op}; (ii) adaptive PF with full braking maneuver (denoted as APF with FB) \cite{Lu2020-fm}; (iii) adaptive PF without the lower road PF when sensing the emergency (denoted as APF without LR); and (iv) {ESPP}-based adaptive PF (denoted as {ESPP}-based APF). {The overall simulation was conducted on a laboratory laptop  with Intel(R) Core(TM) i9-10980HK CPU@2.40GHz and  RAM 32GB.}

\subsection{Environment Settings}

To simulate the high-speed scene, we initialized the longitudinal speed of the adjacent obstacle at 115.2 km/h, and the longitudinal speed of the ego vehicle was initially set at 108 km/h. The adjacent obstacle is suddenly steered without a pre-warning when it is one body position ahead of the ego vehicle to simulate an extreme emergency scene. {For longitudinal control, we assume that the ego vehicle conducts a full brake with a maximum deceleration after the reaction time. Therefore, we directly apply the anti-lock brake system (ABS) controller in the CarSim Simulator when the ``blind alley'' problem is detected.} In addition, we applied five different vehicle mechanical models with 27 degrees of freedom (DOF) \cite{Yin2020-mf} provided in CarSim to simulate different vehicle dynamics, including Sedan, full-size SUV, large Van, full-size pickup, and cargo trucks. The 3D model of the {ESPP} is shown in Fig. \ref{3D_pf_eer}, where we can observe that the {ESPP}-based APF can extend an emergent PF from the road PF for emergency navigation. The emergent PF restricts the ego vehicle from driving within the planned route and achieves a safe stop. {The parameter settings of ESPP-based APF are described in Table. \ref{table1}. It is worth noting that $A_{obs}$ should be set to an immense value, which ensures the obstacle PF is not traversable. Besides, the numerical value of $A_{lane}$ should be small because the vehicle needs to violate the lane PF when a lane-change decision has been made. The numerical values of $\omega_{max}$ and $R_{min}$ are determined based on the selected car model in the CarSim simulator; for example, the D-class sedan is chosen in this study. In addition, $l_w$, $l_f$, and $l_r$ can also be available from the CarSim simulator.} The parameter settings of the MPC controller are denoted in Table. \ref{table2}.

\subsection{Simulation Results}
\begin{table}[t]
    \centering
    \caption{Parameters of ESPP-based PF}
    \label{table1}
    \setlength{\tabcolsep}{16pt}
    \begin{tabular}{cccc}
        \hline
        {$Parm.$} & {$Val.$} &{$Parm.$} & {$Val.$}\\
        \hline
        {$A_{lane}$}&20 &{$A_{obs}$}&150 \\
        {$l_w$} &1.6 m &{$L$} &0.3\\
        {$\zeta$} &1 &{$Y_c$} &4 \\
        {$\upsilon$} &0.75 &$\xi$
         &0.2\\
        {$\eta$} &3 &{$e_{y}^{min,max}$} &$\pm$ 0.7 m \\
        {$l_f$} &1.232 m &$e_{\psi}^{min,max}$ & $\pm$ 0.05 rad\\
        {$Y_{u,l}$} &(8,0) &{$\omega_{max}$} &4.9 \\
        {$l_r$} &1.468 m &$R_{min}$ & 6.12 m\\
        \hline
    \end{tabular}
\end{table}
\begin{table}[t]
    \centering
    \caption{Parameters of MPC controller}
    \label{table2}
    \setlength{\tabcolsep}{15pt}
    \begin{tabular}{cccc}
        \hline
        {$Parm.$} & {$Val.$} & {$Parm.$} 	&{$Val.$}\\
        \hline
        {$N_p$}	&20 &{$N_c$} &5\\ 	
        {$T_s$} &10 [ms] &{$u_{max}$} &0.2 [rad]\\ 	
        {$\Delta u_{max}$} &0.015 [rad] &{$\psi_{max}$} &0.4 [rad]\\
        {$V$} &108 [km/h] &{$\lambda$} &0.15\\ 
        {Q} &$\begin{bmatrix}0.01& 0\\ 0& 0.001 \end{bmatrix}$ &{R} &0.1\\
        \hline
    \end{tabular}
\end{table}
\begin{table*}[t]
    \centering
    \caption{Performance Evaluations}
    \scalebox{0.97}{ 
    \begin{threeparttable}
        \begin{tabular}{@{}ccccccccccccccccc@{}}
        \toprule
        \multicolumn{1}{l}{\multirow{2}{*}{}} & \multicolumn{4}{c}{20 m/s}                              & \multicolumn{4}{c}{25 m/s}                              & \multicolumn{4}{c}{30 m/s}    &   \multicolumn{4}{c}{35 m/s}                \\ \cmidrule(l){2-5} \cmidrule(l){6-9} \cmidrule(l){10-13} \cmidrule(l){14-17} 
        \multicolumn{1}{l}{Planners}                  & AC (1/m)        & RT (s)        & CA        & SS        & AC   (1/m)   & RT (s)        & CA        & SS        & AC (1/m)             & RT (s)        & CA & SS & AC  (1/m)    & RT (s)        & CA        & SS       \\ \midrule
        CPF w CS                              & 5.2e-3          & 5.73          &   \ding{55}    &    \ding{55} & 6.3e-4 & 4.07   &   \ding{55}    &    \ding{55}  & 2.5e-3          & 4.01      &   \ding{55}   &  \ding{55}   & 1.2e-3          & 4.05    &  \ding{55}  &    \ding{55}      \\
        APF w FB                              & 2.6e-3          & 6.18      &  \ding{55}    &  \ding{55}   & 1.4e-3          & 5.55    &   \ding{55}   &  \ding{55}  & 1.1e-3          & 4.21   &  \ding{55}  &   \ding{55} & 7.2e-4 & 4.11 &   \ding{55}    &    \ding{55}  \\
        APF w/o LR       & 4.3e-2          & 6.95    &    \Checkmark     &    \Checkmark   & 3.0e-1          & 6.64          &     \Checkmark      &     \Checkmark      & 3.2e-1          & 7.54    &  \Checkmark  &    \Checkmark  & 3.7e-3 & 3.85 &   \ding{55}    &    \ding{55}   \\ \midrule
        \textbf{ESPP-APF}                     & \textbf{3.7e-3} & \textbf{6.87} & \Checkmark & \Checkmark & \textbf{4.4e-3} & \textbf{6.49} & \Checkmark & \Checkmark & \textbf{1.3e-3} & \textbf{7.25} & \Checkmark  & \Checkmark & \textbf{2.8e-3} & \textbf{8.11} &   \Checkmark    &    \Checkmark  \\ \bottomrule
        \end{tabular}
        \begin{tablenotes}
            \footnotesize
            \item[*] AC: Average Curvature \quad RT: Response Time \quad CA: Collision Avoidance \quad SS: Safe Stop
        \end{tablenotes}
    \end{threeparttable}
    }
\label{per_eva}
\end{table*}

{In Table. \ref{per_eva}, we can observe the overall performance of the four planners under different high speeds. Although both the CPF with CS and APF with FB can produce a smoother path, their response times are relatively shorter than the APF without LR and ESPP-based APF due to the ``blind alley'' problem. Consequently, they fail to accomplish collision avoidance as well as the safe stop for all scenarios. On the other hand, the APF without LR can achieve collision avoidance and safe stop for the scenarios of 20 m/s, 25 m/s, and 30 m/s, but its average curvature is larger than other planners, which means the smoothness of the path and the ride comfort performs worse. Besides, we found the response time of the APF without LR is similar to that of the ESPP-based APF but slightly larger because the planner will stop its response when the vehicle is stopped, which implicitly indicates the APF without LR takes longer to complete the safe stop. In the scenario of 35 m/s, the APF without LR fails to accomplish collision avoidance and has the shortest response time than other planners because it quickly hits the boundary of the emergency stopping lane due to the entire loss of control. On the opposite, the ESPP-based APF can finish all the tasks and produce a smoother path. Next, we will compare and analyze more detailed data.} The trajectories of the ego vehicle obtained by applying the 4 different path planners are shown in Fig. \ref{path}. We can observe that the trajectory of the CPF with CS (denoted by the red dash-dotted line) experiences a side collision with the obstacle (blue vehicle), while that of the APF with FB (denoted by the cyan dashed line) ends up with a rear-end collision with the obstacle. On the other hand, the trajectory of the APF without LR (denoted by the magenta dotted line) successfully avoids the obstacle; however, experiences an apparent tortuous part from 200--160 m. The trajectory of the {ESPP}-based APF also avoids the obstacle while ensuring a smoother curvature in the stopping maneuver. In addition, the total length of the {ESPP}-based APF trajectory was shorter than that of the APF without LR, resulting in a faster-stopping maneuver. 
\begin{figure*}[t]
    \centering
    \includegraphics[width=\hsize]{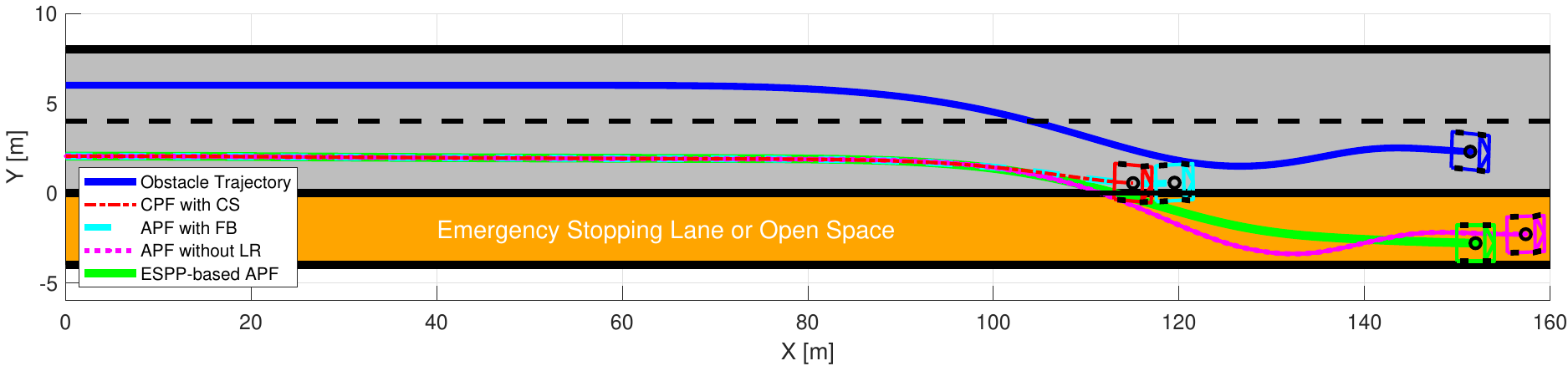}
    \caption{Vehicle trajectories of different path planners: the gray area represents the typical freeway, and the orange area denotes the emergency stopping lane or open space.}
    \label{path}
\end{figure*}
\begin{figure*}[t]
    \centering
    \includegraphics[width=\hsize]{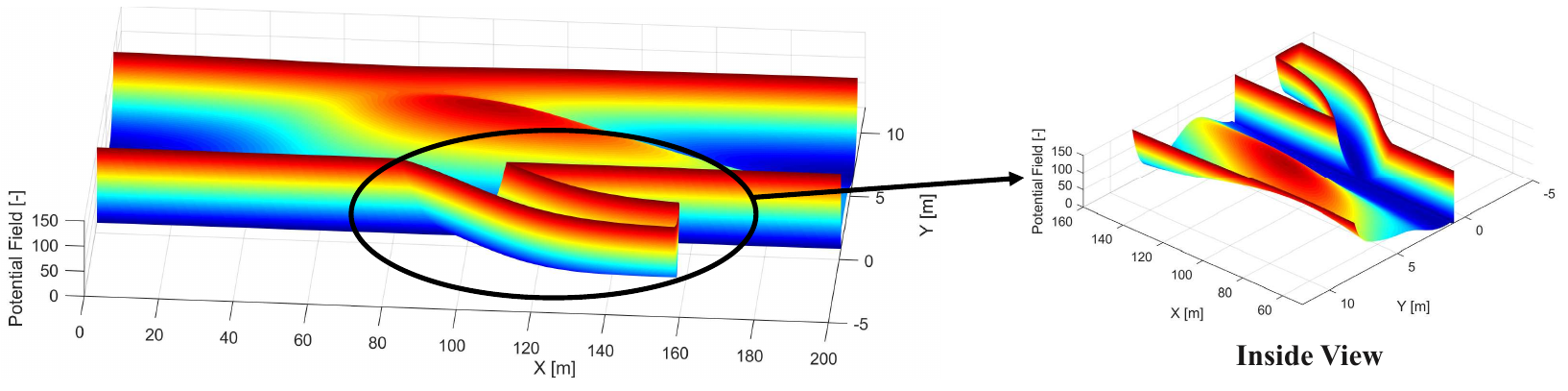}
    \caption{3D potential field modeling of the {ESPP: Opening a breach for the ego-vehicle to avoid collision}}
    \label{3D_pf_eer}
\end{figure*}
\begin{figure*}[t]
\centering
    \begin{minipage}[t]{0.48\textwidth}
        \includegraphics[width=\hsize]{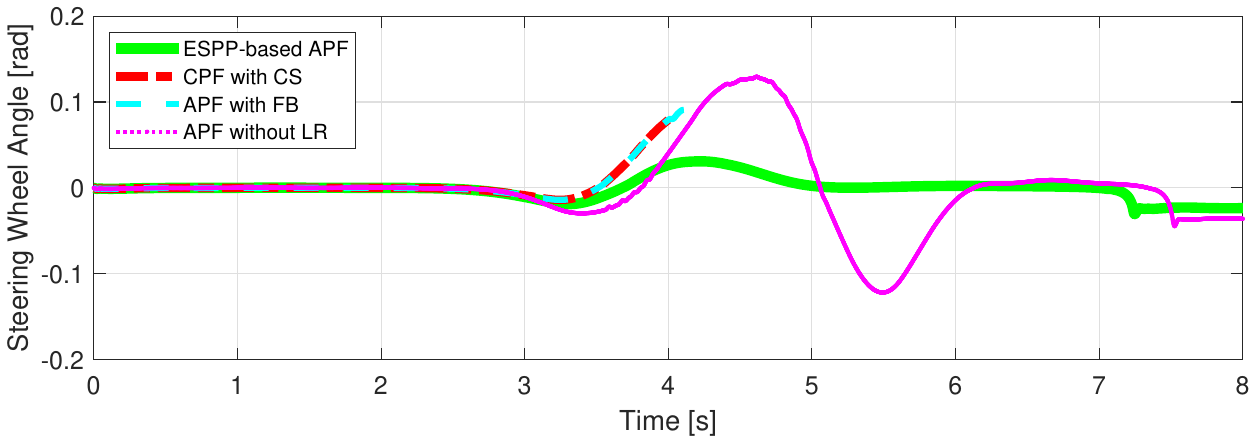}
        \caption{Front wheel steering angles of the ego vehicle.}
        \label{tire_steer}
    \end{minipage}\hspace{0.15in}
    \begin{minipage}[t]{0.48\textwidth}
        \includegraphics[width=\hsize]{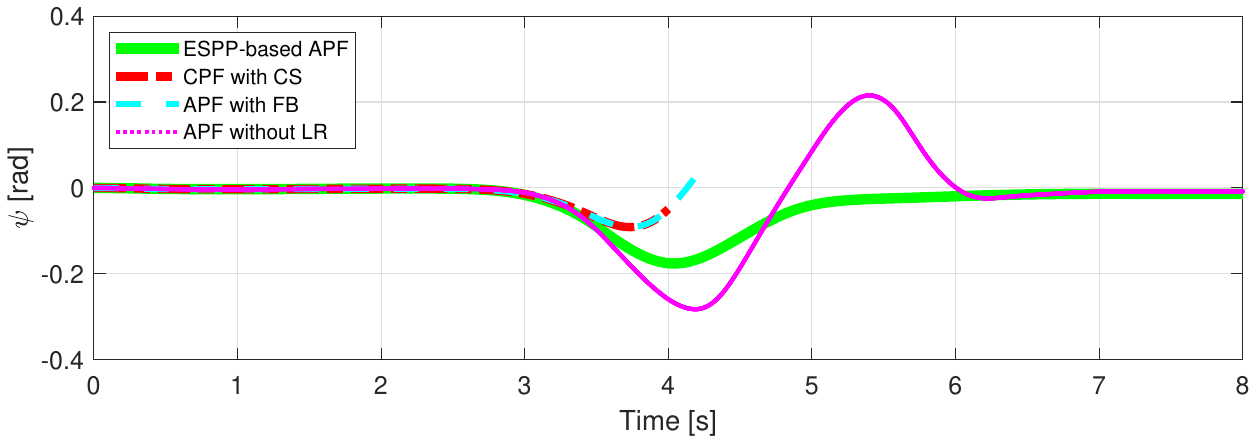}
        \caption{Heading angles of the ego vehicle.}
        \label{psi}
    \end{minipage}
\end{figure*}

As depicted in Fig. \ref{tire_steer}, the front wheel steering angles of the CPF with CS and the APF with FB end up at the time 3.96 s and 4.2 s, respectively, owing to the collision. When the front wheel steering angle of the APF without LR reaches the maximum at 0.13 rad and the minimum at -0.12 rad, during the emergency collision avoidance from 2.67--6.18 s. On the contrary, the front wheel steering angle of the {ESPP}-based APF varies between 0.03 rad and -0.02 rad, which is smaller than that of the APF without LR. Correspondingly, the heading angle of the APF without LR reaches -0.28 and 0.22 rad, which are more extensive than that of the {ESPP}-based APF (with the minimum value at -0.18 rad), as depicted in Fig. \ref{psi}. The longitudinal speeds of the four wheels of the ego vehicle are shown in Fig. \ref{wheel_speed_withoutpf} and \ref{wheel_speed_eer}. We observed that the four wheels of the APF without LR exhibited severe oscillations from 3--5.2 s, which is caused by the excessive wheel slip, requiring a longer time to complete the entire braking maneuver. In contrast, the four wheels of the {ESPP}-based APF exhibited a marginal vibration from 3.2--3.8 [s] that enabled the ego vehicle to complete the braking procedure earlier than the APF without LR. The lateral accelerations of the ego vehicle are shown in Fig. \ref{ay_apf}; the APF without LR exhibited a larger lateral acceleration (up to 0.9 m/s$^2$) than the {ESPP}-based APF (up to -0.48 m/s$^2$) during the emergency collision avoidance. Fig. \ref{stop point} shows the variation of the stop point that is calculated from Algorithm \ref{alg1}. The stop point's X- and Y-axes can jump first at 3.56 s when the emergency situation occurs, and the X-axis varies two times at 4.33 and 6.28 s, respectively, because of the preset constraint conditions. The coordinates of the stop point are finally initialized to $(0,0)$ when the ego vehicle stops. {In addition, Figs. \ref{tire_steer}, \ref{psi}, and \ref{ay_apf} show smoother and smaller responses and control inputs of the system with the proposed ESPP method compared to other methods.}

{From Fig. \ref{tire_roll_rm_withoutPF} to Fig. \ref{wheel_roll_rate_EER}, the rotational dynamics can be observed, including the rolling resist moment (RRM) of four tires and the roll rate (RR) of four wheels. In Fig. \ref{tire_roll_rm_withoutPF}, we can see that the RRMs of Tire L1 and Tire R1 are over 100 N-m in APF without LR-based path planner, while that of ESPP-based path planner maintains between 53 N-m to 102 N-m from 3.2 s to 7.2 s, as depicted in Fig. \ref{tire_roll_rm_EER}. Besides, the RRM of Tire L1 of the APF without LR-based path planner drops dramatically from 104.3 N-m to 13.21 N-m at $T=3.8$ s, possibly due to the tire slip. In addition, the RRMs of the rear tires from both planners vary around 0 N-m. On the other hand, the roll rates of four wheels from the APF without LR-based path planner have obvious oscillations during the braking period (from 3.2 s to 6.7 s) that reaches 10.1 deg/s at $T=4.5$ s and -11.6 deg/s at $T=5.8$ s, as shown in Fig. \ref{wheel_roll_rate_withoutPF}. Conversely, as stated in Fig. \ref{wheel_roll_rate_EER}, the roll rates of four wheels from the ESPP-based path planner vary under $\pm 3.8$ deg/s with a shorter oscillation period.} 

{The computational time is depicted in Fig. \ref{comp_t}, we can observe that the initial computational time of ESPP-based APF is higher than other planners because it has more parameters to be initialized. From $T=3.2$ s to $T=4.5$ s, we can see that the computational time of ESPP-based APF is also increasing more rapidly than other planners due to the activation of the ESPP. In addition, more peaks exist in the green solid line than in other lines after $T=3$ s because the emergency triggering estimation detects the ``blind alley'' problem and activates the ESPP to reach a safe stop, involving solving several constrained optimizations at those peaks. Although the average computational time of the proposed method is higher than other methods, the overall performance is still under 0.04 s, which has conformed to the real-time requirement (within 100 ms) in autonomous driving \cite{jianfeng2023-ad}.}
\begin{figure*}[t]
\centering
\begin{minipage}[t]{0.48\textwidth}
\centering
\includegraphics[width=\hsize]{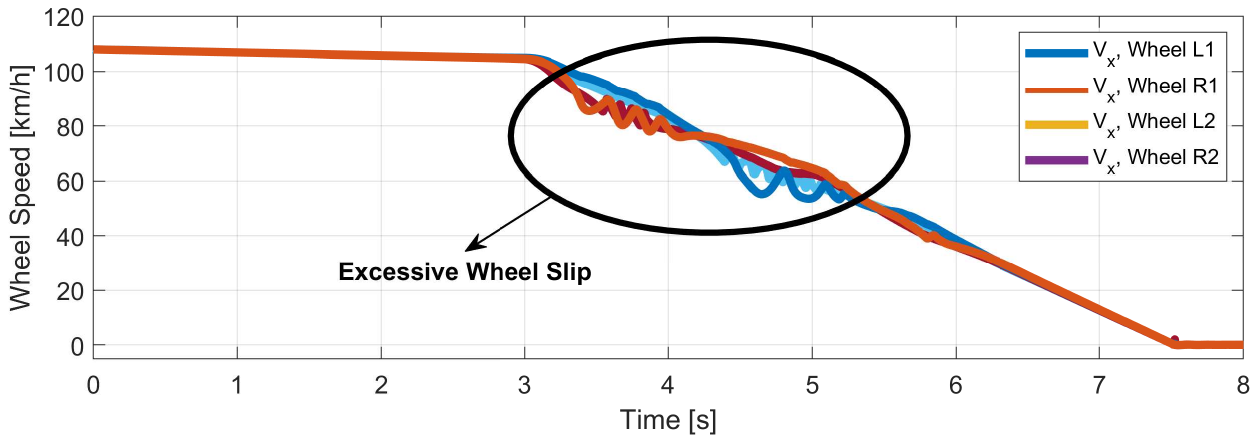}
\caption{{(Linear) Longitudinal speeds} of four wheels of APF without LR: L1 is the left front wheel; R1 is the right front wheel; L2 is the left rear wheel; R2 is the right rear wheel}
\label{wheel_speed_withoutpf}
\end{minipage}\hspace{.15in}
\begin{minipage}[t]{0.48\textwidth}
\centering
\includegraphics[width=\hsize]{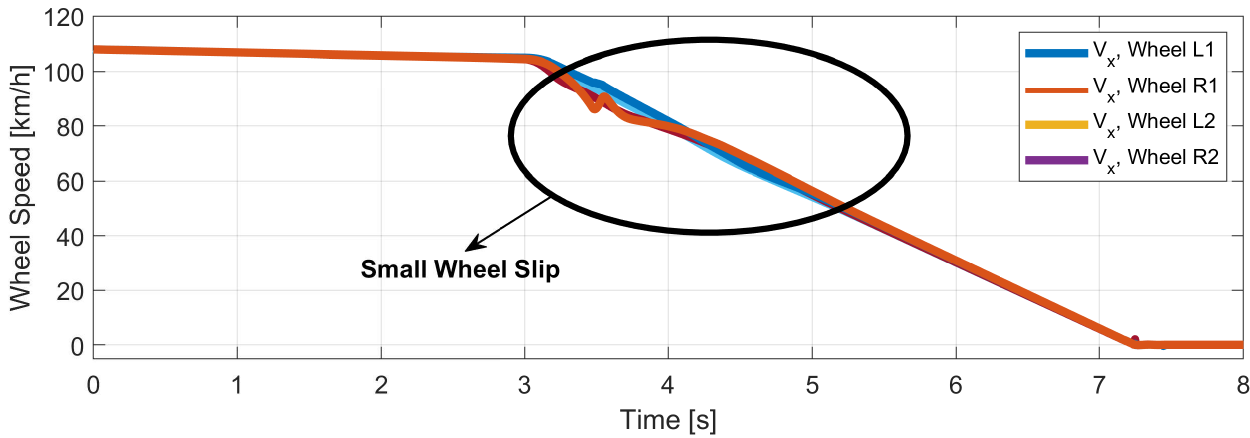}
\caption{{(Linear) Longitudinal speeds} of four wheels of {ESPP}-based APF: L1 is the left front wheel; R1 is the right front wheel; L2 is the left rear wheel; R2 is the right rear wheel}
\label{wheel_speed_eer}
\end{minipage}
\end{figure*}

%
\begin{figure*}[t]
\centering
    \begin{minipage}[t]{0.48\textwidth}
        \includegraphics[width=\hsize]{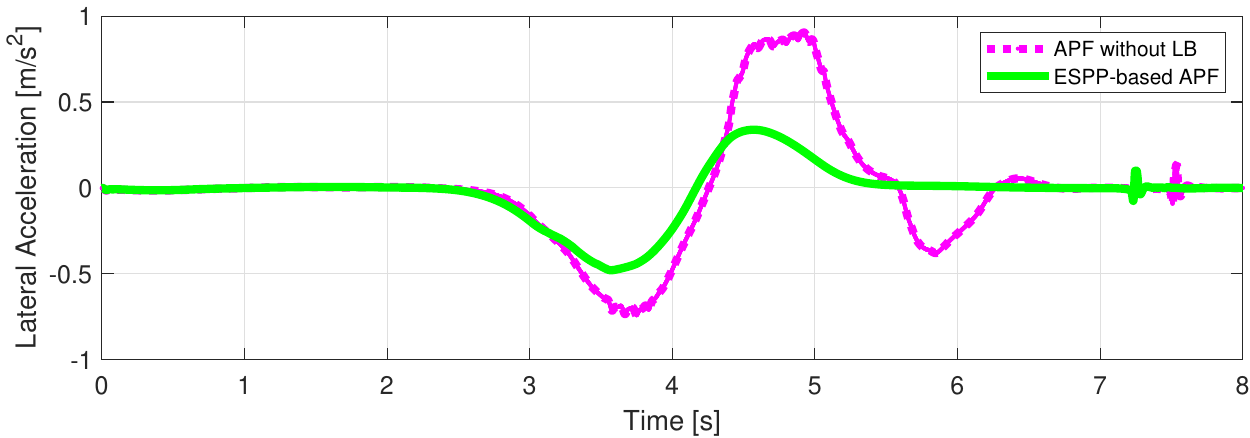}
        \caption{Lateral accelerations of the ego vehicle.}
        \label{ay_apf}
    \end{minipage}\hspace{0.15in}
    \begin{minipage}[t]{0.48\textwidth}
        \includegraphics[width=\hsize]{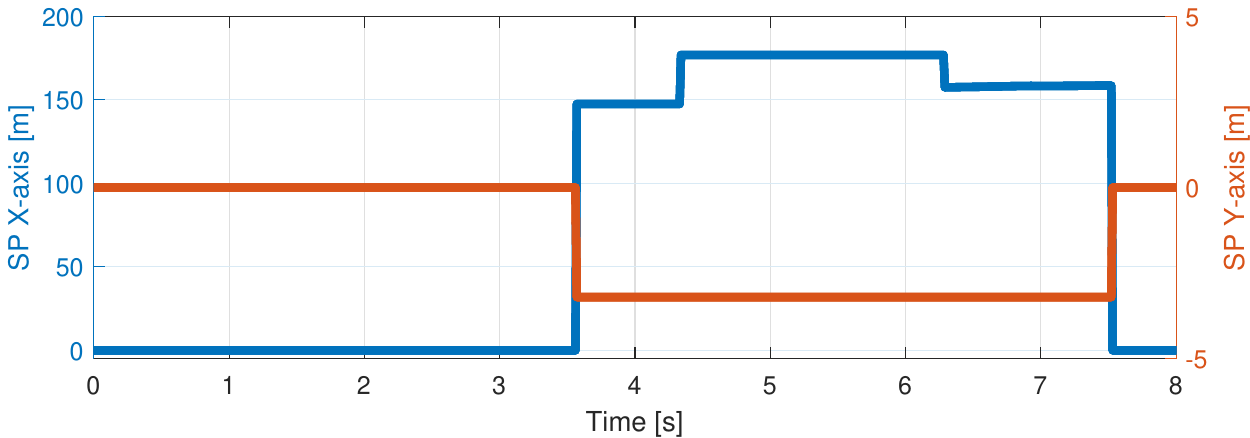}
        \caption{Longitudinal and lateral coordinates of the stop point (SP).}
        \label{stop point}
    \end{minipage}
\end{figure*}

\section{CONCLUSIONS and DISCUSSIONS\label{conclusion}}

In this study, we proposed an {ESPP}-based APF combined with a clothoid curve to overcome an extreme emergency situation in which the obstacle steers recklessly without a pre-warning.  We compared our proposed method with three other path planners in a co-simulation study using MATLAB/Simulink and CarSim simulator to verify its performance. {The simulation results revealed that the proposed method demonstrates effective emergency collision avoidance capabilities, enabling the vehicle to come to a safer stop compared to conventional methods. Moreover, the proposed approach exhibits the advantage of generating a smoother trajectory, contributing to improved ride comfort for passengers. Despite its longer response time when compared to other methods, even in high-speed scenarios, the proposed method still exhibits reliable collision avoidance performance.
Additionally, the proposed method addresses the issue of wheel slip, a phenomenon that typically occurs during sudden braking, and effectively mitigates its effects. By doing so, it ensures a more stable and controlled deceleration, enhancing overall safety during critical braking maneuvers.
Furthermore, the benefits of the proposed method extend beyond safety considerations, as it also generates a more comfortable lateral acceleration. This improvement in lateral dynamics contributes to a smoother and more pleasant ride experience for vehicle occupants.
Taken together, the simulation results demonstrate the effectiveness and versatility of the proposed method, making it a promising solution for enhancing both safety and ride quality in emergency scenarios.}

{In this study, we have an underlying assumption that the ego vehicle possesses complete knowledge of obstacle information. However, this assumption could result in failures when encountering occlusion problems, where certain obstacles might not be fully visible or detectable. Additionally, we need to further verify the robustness of the proposed algorithm against parameter changes to avoid potential failures under varying conditions.
Regarding the current limitations, our investigation has been limited to simulations on straight roads, which poses a foreseeable restriction on the applicability of the proposed approach to other road types, such as curved roads and intersections. Expanding our research to include different road scenarios would provide a more comprehensive evaluation of the algorithm's effectiveness and practicality.
Furthermore, we employed a 2-degree-of-freedom (2-DOF) vehicle model in the MPC controller, which may require enhancements to represent real-world driving conditions accurately. For instance, paying attention to the road bank angle might lead to suboptimal control decisions. To address this, we intend to explore a higher degree of freedom vehicle dynamics model for the tracking controller in future work. However, a higher dimensional vehicle model will increase computational costs that should be considered based on specific application scenarios and performance requirements. Adopting a more sophisticated vehicle model could yield different outcomes when integrating it with the proposed ESPP method.}
\begin{figure*}[t]
\centering
    \begin{minipage}[t]{0.48\textwidth}
        \includegraphics[width=\hsize]{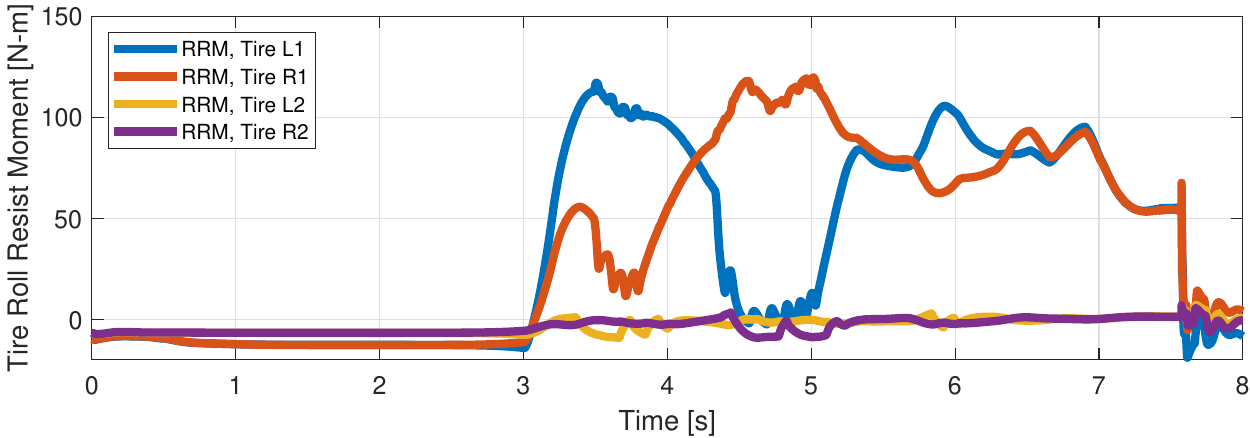}
        \caption{{Rolling resist moment of four tires from the APF without LR-based planner: L1 is the left front tire; R1 is the right front tire; L2 is the left rear tire; R2 is the right rear tire}}
        \label{tire_roll_rm_withoutPF}
    \end{minipage}\hspace{0.15in}
    \begin{minipage}[t]{0.48\textwidth}
        \includegraphics[width=\hsize]{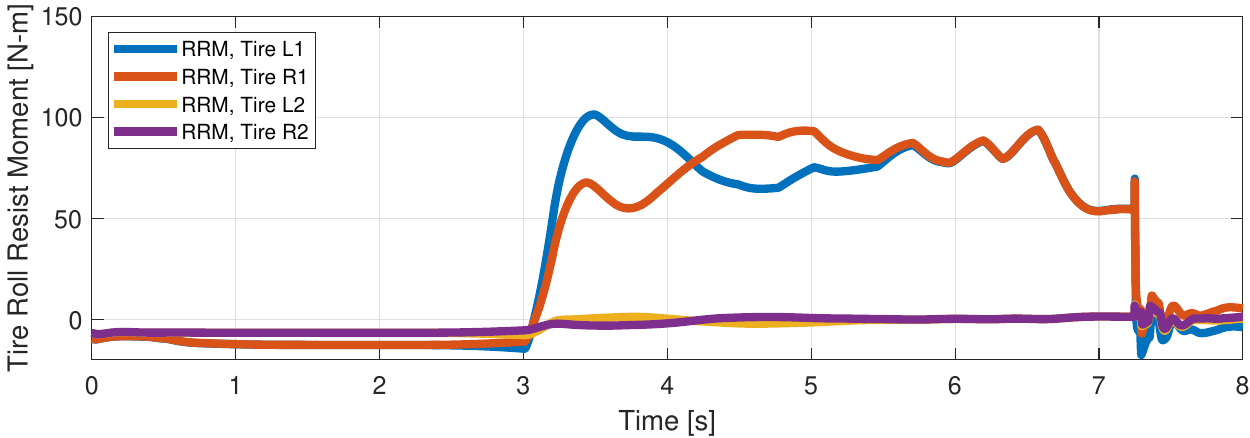}
        \caption{{Rolling resist moment of four tires from the ESPP-based planner: L1 is the left front tire; R1 is the right front tire; L2 is the left rear tire; R2 is the right rear tire}}
        \label{tire_roll_rm_EER}
    \end{minipage}
\end{figure*}
\begin{figure*}[t]
\centering
    \begin{minipage}[t]{0.48\textwidth}
        \includegraphics[width=\hsize]{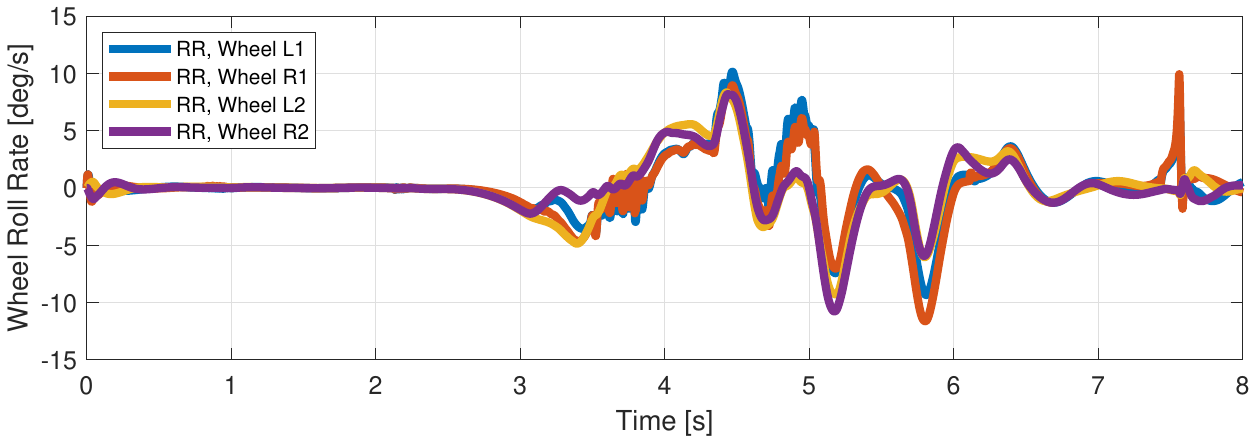}
        \caption{{Roll rate of four wheels from the APF without LR-based planner}}
        \label{wheel_roll_rate_withoutPF}
    \end{minipage}\hspace{0.15in}
    \begin{minipage}[t]{0.48\textwidth}
        \includegraphics[width=\hsize]{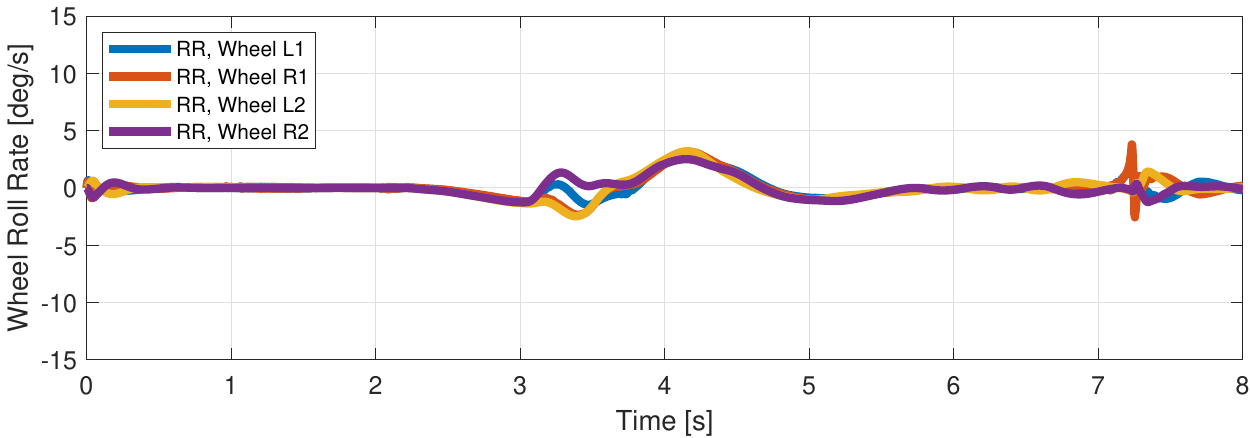}
        \caption{{Roll rate of four wheels from the ESPP-based planner}}
        \label{wheel_roll_rate_EER}
    \end{minipage}
\end{figure*}
\begin{figure}[t]
    \centering
    \includegraphics[width=\hsize]{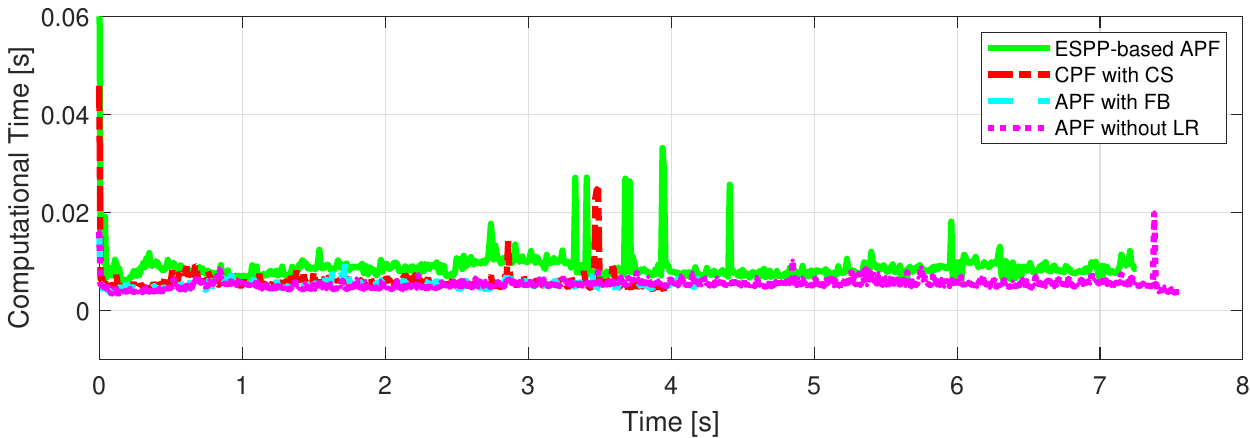}
    \caption{{Computational time of the planners.}}
    \label{comp_t}
\end{figure}

\section*{Acknowledgment}

{We extend our heartfelt gratitude to Dr. Maxime Clement from Tier IV Inc. for his invaluable insights, guidance, and thought-provoking suggestions throughout the course of this research.} These research results were partly sponsored by the China Scholarship Council (CSC) program (No.202208050036) and the Japan Society for the Promotion of Science (JSPS) Research Fellowship for Young Scientists (DC2) program (grant number: 23KJ0391).

\bibliographystyle{IEEEtran}
\bibliography{reference}

\vfill

\end{document}